\documentclass{article}

\PassOptionsToPackage{numbers, compress}{natbib}
\usepackage[preprint]{neurips_2022}

\usepackage[utf8]{inputenc} % allow utf-8 input
\usepackage[T1]{fontenc}    % use 8-bit T1 fonts
\usepackage{hyperref}       % hyperlinks
\usepackage{url}            % simple URL typesetting
\usepackage{booktabs}       % professional-quality tables
\usepackage{amsfonts}       % blackboard math symbols
\usepackage{nicefrac}       % compact symbols for 1/2, etc.
\usepackage{microtype}      % microtypography
\usepackage{xcolor}         % colors

\usepackage{graphicx}
\usepackage{amsmath}
\usepackage{amssymb}
\usepackage{adjustbox,lipsum}
\usepackage{kotex}
\usepackage{multirow}

\title{MSSNet: Multi-Scale-Stage Network for Single Image Deblurring}

\author{
    \vspace*{-7pt}\\
    Kiyeon Kim
    \qquad
    Seungyong Lee
    \qquad
    Sunghyun Cho\\
    \vspace*{-7pt}\\
    POSTECH\\
    \vspace*{-7pt}\\
    {\tt\small
    \{kiyeon, leesy, s.cho\}@postech.ac.kr}\\
}

\begin{document}

\maketitle
\newcommand{\Eq}[1]  {Eq.\ (\ref{eq:#1})}
\newcommand{\Eqs}[1] {Eqs.\ (\ref{eq:#1})}
\newcommand{\Fig}[1] {Fig.\ \ref{fig:#1}}
\newcommand{\Figs}[1]{Figs.\ \ref{fig:#1}}
\newcommand{\Tbl}[1]  {Table \ref{tbl:#1}}
\newcommand{\Tbls}[1] {Tables \ref{tbl:#1}}
\newcommand{\Sec}[1] {Sec.\ \ref{sec:#1}}
\newcommand{\Secs}[1] {Secs.\ \ref{sec:#1}}
\newcommand{\etal}   {{\textit{et al.}}}

\newcommand{\setone}[1] {\left\{ #1 \right\}} % math set notation { a }
\newcommand{\settwo}[2] {\left\{ #1 \mid #2 \right\}} % math set notation { a | b}

\definecolor{brown}{rgb}{0.65, 0.16, 0.16}
\definecolor{purp}{rgb}{0.65, 0.16, 0.65}
\definecolor{orange}{rgb}{1.0, 0.5, 0.0}
\definecolor{blue}{rgb}{0.0, 0.5, 1.0}
\definecolor{green}{rgb}{0, 0.8, 0}
\definecolor{lgreen}{rgb}{0.6, 0.8, 0}
\definecolor{red}{rgb}{0.8, 0, 0}
\definecolor{darkblue}{rgb}{0, 0.2, 0.6}

\newcommand{\todo}[1]{{\textcolor{blue}{TODO: #1}}}
\newcommand{\son}[1]{{\textcolor{magenta}{lee: #1}}}
\newcommand{\sean}[1]{{\textcolor{green}{sean: #1}}}
\newcommand{\sunghyun}[1]{{\textcolor[rgb]{0.6,0.0,0.6}{sunghyun: #1}}}
\newcommand{\rjs}[1]{{\textcolor[rgb]{1,0.0,0.0}{rjs: #1}}}
\newcommand{\kyoungkook}[1]{{\textcolor[rgb]{0.6,0.6,1}{kyoungkook: #1}}}
\newcommand{\change}[1]{{\color{red}#1}}

% color (command added by ky)
\newcommand{\blue}[1]{{\textcolor[rgb]{0,0,0.9}{#1}}}
\newcommand{\red}[1]{{\textcolor[rgb]{0.9,0,0}{#1}}}
\newcommand{\violet}[1]{{\textcolor[rgb]{0.4,0,0.6}{#1}}}
\newcommand{\green}[1]{{\textcolor[rgb]{0,0.5,0}{#1}}}

% LaTeX commands to reduce the spacing above and below figures
\renewcommand{\topfraction}{0.95}
\setcounter{bottomnumber}{1}
\renewcommand{\bottomfraction}{0.95}
\setcounter{totalnumber}{3}
\renewcommand{\textfraction}{0.05}
\renewcommand{\floatpagefraction}{0.95}
\setcounter{dbltopnumber}{2}
\renewcommand{\dbltopfraction}{0.95}
\renewcommand{\dblfloatpagefraction}{0.95}

\newcommand{\argmin}{\mathop{\mathrm{argmin}}\limits} 

\renewcommand{\paragraph}[1]{{\vspace{0em}\noindent\textbf{#1.}}}

%% usually not used, instead use the ordinary latex comment %
%\newcommand{\comment}[1]{} 

\begin{abstract}
   Most of traditional single image deblurring methods before deep learning adopt a coarse-to-fine scheme that estimates a sharp image at a coarse scale and progressively refines it at finer scales. While this scheme has also been adopted to several deep learning-based approaches, recently a number of single-scale approaches have been introduced showing superior performance to previous coarse-to-fine approaches both in quality and computation time. In this paper, we revisit the coarse-to-fine scheme, and analyze defects of previous coarse-to-fine approaches that degrade their performance. Based on the analysis, we propose Multi-Scale-Stage Network (MSSNet), a novel deep learning-based approach to single image deblurring that adopts our remedies to the defects. Specifically, MSSNet adopts three novel technical components: stage configuration reflecting blur scales, an inter-scale information propagation scheme, and a pixel-shuffle-based multi-scale scheme. Our experiments show that MSSNet achieves the state-of-the-art performance in terms of quality, network size, and computation time.
\end{abstract}
\section{Introduction}
\label{sec:intro}

Single image deblurring aims to restore a sharp image from a blurry one caused by camera shake or object motion. As blur severely degrades the image quality and the performance of other tasks such as object detection, deblurring has been extensively studied for decades~\cite{Fergus:2006:removing,Cho:2009:fast,Levin:2011:efficient,Schuler:2015:learning,Pan:2016:blind,Chakrabarti:2016:neural,Nah:2017:DeepDeblur,Tao:2018:SRN,Zhang:2019:DMPHN,Suin:2020:SAPHN,Zamir:2021:MPRNet,Cho:2021:MIMO}.

% coarse-to-fine scheme of classical approaches
Most of classical single image deblurring approaches before deep learning estimate a blur kernel, which describes how an image has been blurred, and a latent sharp image through alternating optimization~\cite{Fergus:2006:removing,Shan:2008:high,Cho:2009:fast,Levin:2009:understanding,Xu:2010:two,Levin:2011:efficient,Xu:2013:unnatural,Sun:2013:edge,Pan:2016:blind,Cho:2017:convergence}.
For computational efficiency and accuracy in estimating a blur kernel and latent sharp image, a coarse-to-fine scheme has been widely adopted by classical approaches~\cite{Shan:2008:high,Cho:2009:fast,Xu:2010:two,Sun:2013:edge,Xu:2013:unnatural,Cho:2017:convergence}.
The coarse-to-fine scheme estimates a small blur kernel and latent image at a coarse scale and uses them as an initial solution at the next scale.
The small sizes of both images and blur at a coarse scale enable computationally efficient estimation.
Also, the small blur size at a coarse scale enables more accurate estimation of a blur kernel and latent image.
As a result, the coarse-to-fine scheme can quickly provide an accurate initial solution to the next scale, and improve both quality and efficiency of deblurring.

% coarse-to-fine scheme of deep learning approaches
Thanks to the effectiveness of the coarse-to-fine scheme proven by traditional approaches, it has also been adopted to several deep learning-based single image deblurring approaches, such as DeepDeblur~\cite{Nah:2017:DeepDeblur}, SRN~\cite{Tao:2018:SRN}, and PSS-NSC~\cite{Gao:2019:PSS-NSC}.
These approaches directly restore a latent sharp image from a blurry image without blur kernel estimation.
They adopt multi-scale neural network architectures that stack sub-networks for different scales to initially estimate a small-scale latent image and then a large-scale latent image using the small-scale latent image as a guidance.
While they do not estimate blur kernels, they share the same motivation with classical approaches: as the image and blur sizes are small at a coarse scale, a deblurred image can be estimated more efficiently and accurately.

% recent single-scale deep learning approaches
Nonetheless, several deep learning-based single-scale approaches have recently been introduced.
Specifically, Zhang \textit{et al.}~\cite{Zhang:2019:DMPHN} pointed out the expensive computation time of the previous multi-scale approaches and the relatively low contribution of lower scale results on the final deblurring quality, and proposed an alternative single-scale approach named DMPHN.
Following Zhang \textit{et al.}, Suin \textit{et al.}~\cite{Suin:2020:SAPHN} and Zamir \textit{et al.}~\cite{Zamir:2021:MPRNet} also proposed hierarchical multi-stage methods based on DMPHN.
These approaches show superior performance to previous multi-scale approaches both in quality and computation time, making the traditional coarse-to-fine scheme seem obsolete.

% Our contribution
In this paper, we address the following questions.
The motivations of the coarse-to-fine scheme still look valid, but why do the coarse-to-fine approaches perform worse than recent single-scale approaches?
What degrades their performance and how can we fix them?
To this end, we revisit the coarse-to-fine scheme and analyze the defects of previous coarse-to-fine approaches that degrade their performance but have been \emph{overlooked} so far.

Based on the analysis, we propose Multi-Scale-Stage Network (MSSNet), a novel deep learning-based deblurring approach that adopts a coarse-to-fine scheme with our remedies to the defects.
MSSNet consists of multiple scales and multiple stages at each scale.
To remedy the defects of previous coarse-to-fine approaches, MSSNet adopts three novel strategies: stage configuration reflecting blur scales, an inter-scale information propagation scheme, and a pixel-shuffle-based multi-scale scheme.
Each strategy is simple and straightforward, resulting in simple architecture for MSSNet.
Nonetheless, our experiments show that, once the details are done right, this model can achieve state-of-the-art performance in terms of quality, network size, and computation time.
Our simple yet effective architecture can serve as a strong baseline and our strategies can provide a guideline for future deblurring research.

\section{Related Work}
\label{sec:related}
% traditional approaches

Traditional single image deblurring methods ~\cite{Fergus:2006:removing,Shan:2008:high,Cho:2009:fast,Levin:2009:understanding,Xu:2010:two,Levin:2011:efficient,Xu:2013:unnatural,Sun:2013:edge,Pan:2016:blind,Cho:2017:convergence} before deep learning assume blur models that describe how a blurred image is obtained using blur kernels. %, and solve inverse problems based on the models.
Unfortunately, they often fail due to their restrictive blur models and the ill-posedness of the problem.
To improve deblurring quality, convolutional neural networks (CNNs) have recently been adopted~\cite{Sun:2015:learning,Hradivs:2015:convolutional,Chakrabarti:2016:neural,Schuler:2015:learning}.
For example, Schuler \etal~\cite{Schuler:2015:learning} and Sun \etal~\cite{Sun:2015:learning} proposed CNNs that estimate blur kernels and a latent image based on traditional blur models.
However, as they still rely on blur models, their performances are limited.
To overcome such limitation, deep learning-based methods that directly restore sharp images without blur kernels have been proposed \cite{Nah:2017:DeepDeblur,Tao:2018:SRN,Zhang:2019:DMPHN,Suin:2020:SAPHN,Zamir:2021:MPRNet,Cho:2021:MIMO}.
These methods can be broadly categorized into single- and multi-scale approaches with respect to their network architectures and training strategies.

\paragraph{Single-Scale Approaches}
Recently, single-scale multi-stage architectures~\cite{Zhang:2019:DMPHN,Suin:2020:SAPHN,Zamir:2021:MPRNet,Chen:2021:HINet} are gaining popularity. Zhang \etal~\cite{Zhang:2019:DMPHN} proposed DMPHN, the first multi-stage network based on a multi-patch approach in single image deblurring. This approach splits an image into multiple disjoint patches and processes each patch independently while gradually merging them in a hierarchical manner through multiple stages. Based on the multi-patch approach, Suin \etal~\cite{Suin:2020:SAPHN} proposed a dynamic filtering module to remove spatially varying blurs. Zamir \etal~\cite{Zamir:2021:MPRNet} proposed MPRNet, which progressively removes blur by giving supervision at each stage.
Chen \etal~\cite{Chen:2021:HINet} introduced half-instance normalization to the multi-stage architecture.
Besides multi-stage architectures, Purohit \etal~\cite{Purohit:2020:RADN} proposed a deep single-stage architecture based on DenseNet~\cite{Huang:2017:DenseNet}.
However, these single-scale approaches do not use initial solutions estimated from coarse scales, so they are less efficient and accurate as will be shown in \Sec{evaluation}.

\paragraph{Multi-Scale Approaches}
Multi-scale approaches are typically based on multi-scale neural network architectures that stack sub-networks in a hierarchical way, and training strategies that train each sub-network to produce deblurred images at different scales.
Nah \etal~\cite{Nah:2017:DeepDeblur} proposed DeepDeblur, the first end-to-end deep learning-based method that adopts a multi-scale neural network to directly restore a latent sharp image from a blurry image in a coarse-to-fine manner.
Each sub-network consists of ResBlocks~\cite{He:2016:ResNet}, and is trained to produce a deblurred image of its corresponding scale.
Tao \etal~\cite{Tao:2018:SRN} presented SRN, which adopts a UNet-based architecture \cite{Ronneberger:2015:UNet} for each scale.
Gao \etal~\cite{Gao:2019:PSS-NSC} also proposed a UNet-based multi-scale architecture with a different parameter sharing strategy.
However, their performance is limited due to the drawbacks of their network architectures as we will discuss in \Sec{coarse_to_fine_analysis}.
Besides these approaches, Cho \etal~\cite{Cho:2021:MIMO} recently proposed MIMO-UNet, which adopts a single UNet \cite{Ronneberger:2015:UNet} with multi-scale loss terms.
This approach is, however, different from a conventional coarse-to-fine approach as it has a large encoder that processes an input image in a fine-to-coarse manner.
Furthermore, as \Sec{evaluation} will show, our MSSNet outperforms MIMO-UNet with much fewer parameters and computations.

\section{Shortcomings of Previous Coarse-to-Fine Approaches}
\label{sec:coarse_to_fine_analysis}

%%%%%%%%%%%%%%%%%%%%%%%%%%%%%%%%
\begin{figure}[t]
    \begin{center}
    \includegraphics[width=0.65\linewidth]{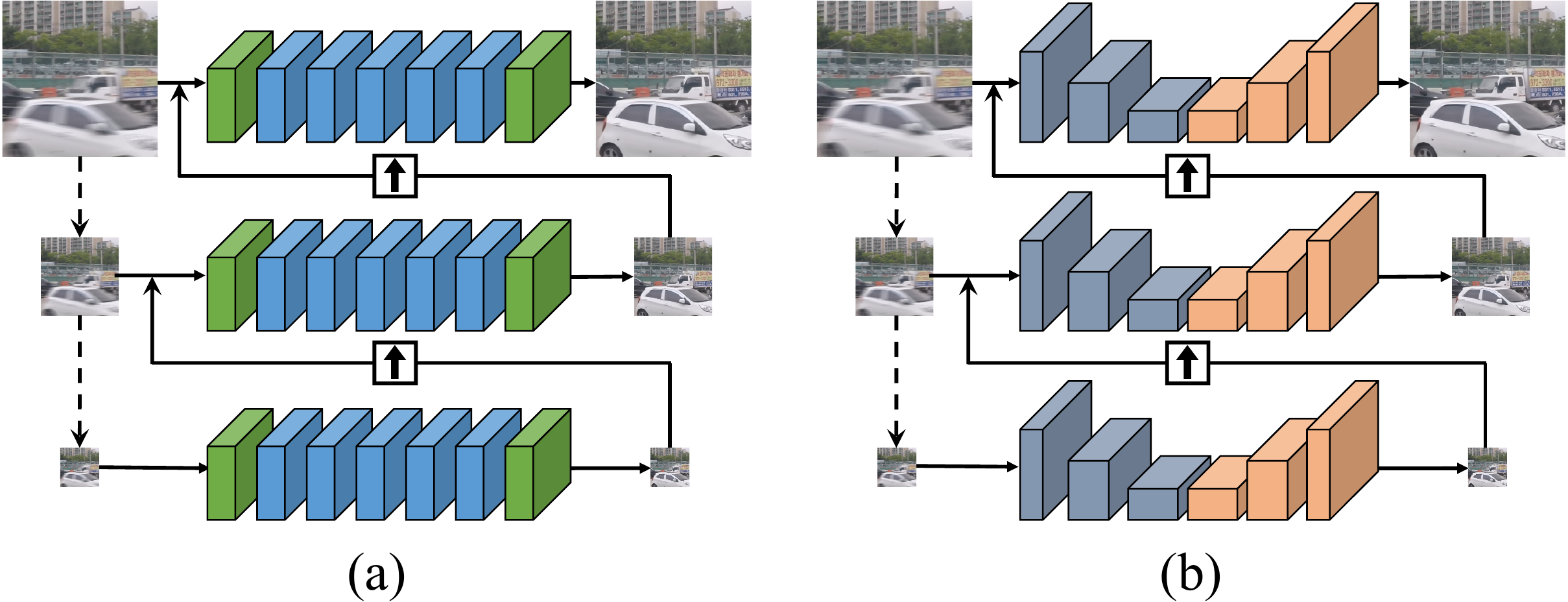}
    \end{center}
    \vspace{-0.4cm}
    \caption{Previous coarse-to-fine architectures. (a) DeepDeblur~\cite{Nah:2017:DeepDeblur}. (b) SRN~\cite{Tao:2018:SRN} and PSS-NSC~\cite{Gao:2019:PSS-NSC}.
}
    \label{fig:previous_coarse_to_fine}
    \vspace{-0.4cm}
\end{figure}
%%%%%%%%%%%%%%%%%%%%%%%%%%%%%%%%%

This section analyzes architectural defects of previous coarse-to-fine approaches, and discusses our ideas to remedy them. MSSNet with our remedies is presented in \Sec{MSSNet}.

\Fig{previous_coarse_to_fine} illustrates the network architectures of previous coarse-to-fine approaches~\cite{Nah:2017:DeepDeblur,Tao:2018:SRN,Gao:2019:PSS-NSC}.
While SRN~\cite{Tao:2018:SRN} adopts additional recurrent connections between consecutive scales to achieve additional performance gain, which is omitted in the figure,
the previous coarse-to-fine approaches share essentially the same deblurring process.
All the methods first build an image pyramid by downsampling an input blurred image.
Then, from the coarsest scale, they estimate a deblurred image from a downsampled blurred image, upsample the deblurred image, and feed it to the sub-network at the next scale.
The sub-network at the next scale then estimates a deblurred image from the blurred image at the current scale using the deblurred image from the previous scale as a guidance.
All the sub-networks at different scales share the same network architecture.
In the following, we analyze the shortcomings of these approaches one by one and present our ideas to address them.

\paragraph{Network architectures disregarding blur scales}
The first shortcoming of the previous approaches is their network architectures that disregard blur scales.
Blur spreads a pixel value in a latent image over an area of the blur size.
Thus, to restore the pixel value at a certain pixel, it is essential to use receptive fields larger than the blur size to aggregate information spread over the area.
Consequently, larger blur sizes require larger receptive fields or deeper neural networks.
Likewise, a coarse-to-fine approach needs deeper sub-networks for finer scales.
While the previous coarse-to-fine approaches use a deblurred image from the previous scale to deblur the blurred image at the current scale \cite{Nah:2017:DeepDeblur,Tao:2018:SRN,Gao:2019:PSS-NSC},
large receptive fields are still required for finer scales.
In multi-scale approaches, a deblurred image from a lower scale lacks fine details as it is estimated from a downsampled image, and such fine details must be restored from the blurred image at a finer scale. Restoring detail at one pixel inevitably needs to aggregate information spread over an area of the blur size regardless of a result from the previous scale.
Thus, it is still more effective to have deeper sub-networks for finer scales as will be shown in our experiments.

\paragraph{Ineffective information propagation across scales}
The previous coarse-to-fine approaches pass the pixel values of a deblurred result from a coarse scale to the next scale \cite{Nah:2017:DeepDeblur,Tao:2018:SRN,Gao:2019:PSS-NSC}.
This causes a significant loss of abundant information encoded in the feature vectors at coarse scales, and eventually degrades the deblurring performance.

\paragraph{Information loss caused by downsampling}
To produce multi-scale input blurred images, the previous approaches build an image pyramid by repeatedly downsampling an input image \cite{Nah:2017:DeepDeblur,Tao:2018:SRN,Gao:2019:PSS-NSC}.
Unfortunately, downsampling causes significant information loss.
Specifically, a downsampling operation reduces the pixels not only in the input image but also in its deblurred result by 1/4, which severely limits the quality of a guidance to the next scale.
To overcome this, in our approach, we present a multi-scale scheme based on the pixel-shuffle~\cite{Shi:2016:ESPCN} operation that reduces the spatial resolution without information loss.
%-------------------------------------------------------------------------

\section{Multi-Scale-Stage Network}
\label{sec:MSSNet}

%%%%%%%%%%%%%%%%%%%%%%%%%%%
\begin{figure}[t]
    \begin{center}
    \includegraphics[width=0.9\linewidth]{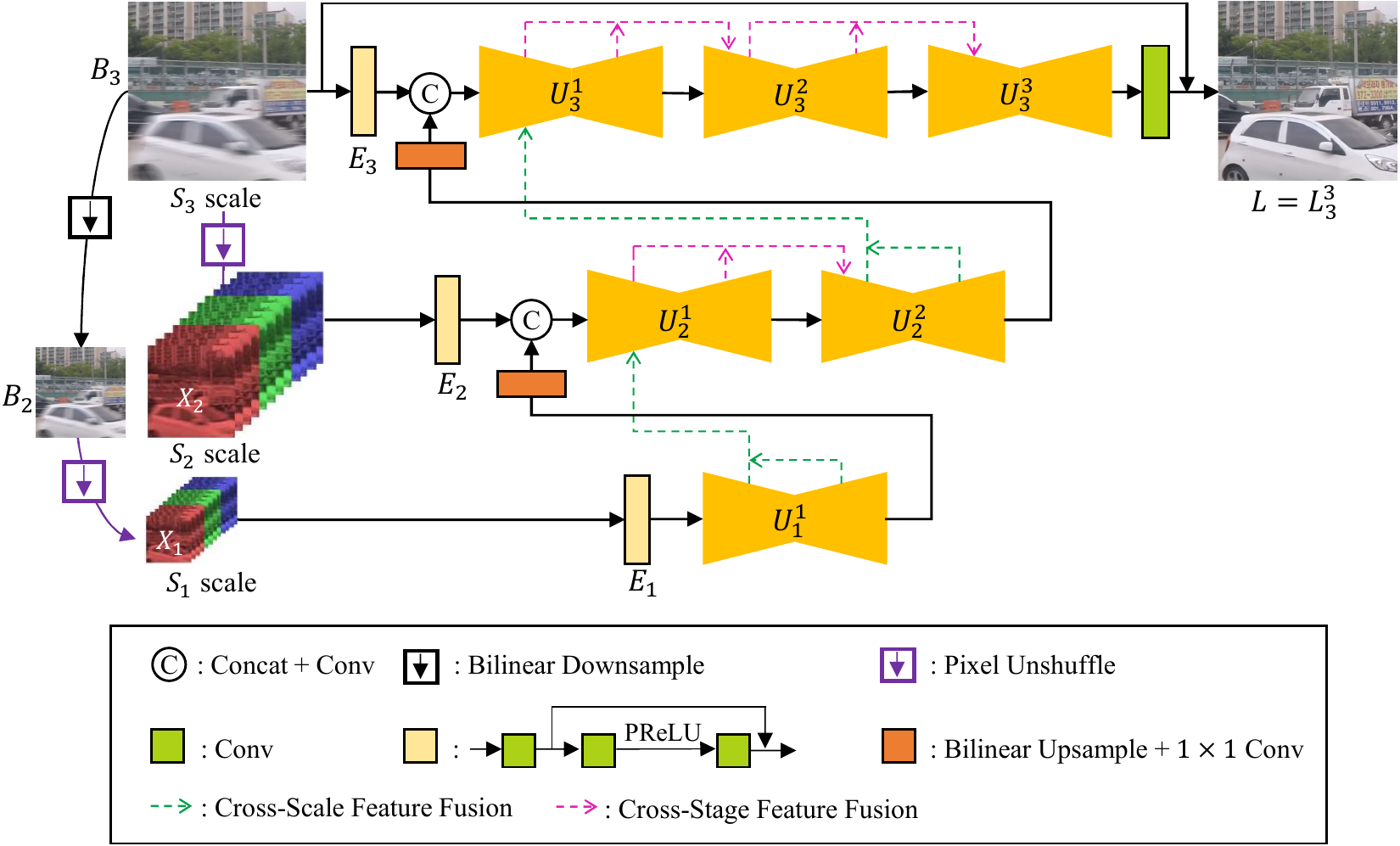}
    \end{center}
    \vspace{-9pt}
    \caption{Network architecture of MSSNet.}
    \label{fig:MSSNet-Test}
\end{figure}
%%%%%%%%%%%%%%%%%%%%%%%%%%%

%%%%%%%%%%%%%%%%%%%%%%%%%%
\begin{figure}[t]
    \begin{center}
    \includegraphics[width=0.8\linewidth]{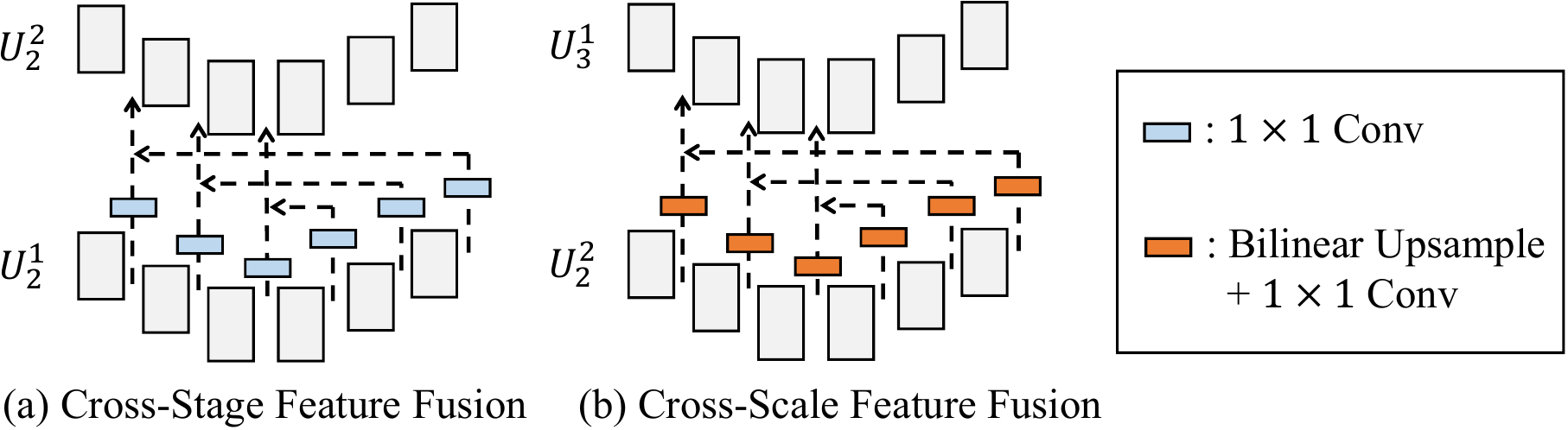}
    \end{center}
    \vspace{-9pt}
    \caption{Cross-stage and cross-scale feature fusion schemes.}
    \label{fig:model_detail}
\end{figure}
%%%%%%%%%%%%%%%%%%%%%%%%%%

\subsection{Network Architecture}

In this section, we present MSSNet, which is designed based on the analysis in \Sec{coarse_to_fine_analysis}.
\Fig{MSSNet-Test} illustrates the architecture of MSSNet.
MSSNet is composed of three scales following previous coarse-to-fine approaches~\cite{Nah:2017:DeepDeblur,Tao:2018:SRN,Gao:2019:PSS-NSC}.
We denote each scale by $S_1$, $S_2$, and $S_3$ from the coarsest to finest scales, respectively.
MSSNet takes a single input blurred image $B$ and estimates a deblurred image $L$ in a coarse-to-fine manner.
For effective restoration, MSSNet adopts the residual learning scheme, which has been widely adopted in various restoration tasks~\cite{Kim:2016:VDSR,Zhang:2017:DnCNN,Lai:2018:LapSRN,Park:2020:MTRNN,Zamir:2021:MPRNet,Cho:2021:MIMO}, i.e.,
MSSNet predicts a residual image $R$, which is added to the input blurred image $B$ to obtain a deblurred output $L=B+R$.
A detailed architecture of MSSNet can be found in the appendix.

MSSNet is specifically designed to reflect blur scales, to facilitate effective inter-scale information propagation, and to avoid information loss caused by downsampling.
We describe each component of our network in the following.

\paragraph{Stage Configuration Reflecting Blur Scales}
To reflect blur scales, the sub-networks of MSSNet at finer scales are designed to have deeper architectures.
Specifically, each scale of MSSNet has one, two and three stages from $S_1$ to $S_3$, respectively, where each stage consists of a single light-weight UNet module~\cite{Ronneberger:2015:UNet}.
We denote each UNet module by $U_{i}^{j}$ where $i$ and $j$ are scale and stage indices, respectively.
The modules share the same network architecture but have different weights.
%The detailed architecture can be found in the supplementary material.
Each module is trained to produce residual features that can be converted to a residual image and added to a blurred image to produce a deblurred image.
More details on the training of MSSNet is explained in \Sec{training}.

\paragraph{Inter-Scale Information Propagation}
Whereas the existing multi-scale networks deliver an upsampled deblurred image from a coarse scale to the next scale as an initial solution, MSSNet delivers upsampled residual features to facilitate effective information propagation between scales. 
Specifically, at the end of a coarse scale, residual features are bilinearly upsampled and processed through a $1\times1$ conv layer.
Then, the resulting features are concatenated to the features from a blurred image at the next scale and convolved with $3\times3$ filters to produce fused features. The fused features are then fed into the UNet modules to produce deblurred residual features at the current scale.

\paragraph{Pixel-Shuffle-Based Multi-Scale Scheme}
To avoid information loss caused by the downsampling operations when producing multi-scale input blurred images, we propose a pixel-shuffle~\cite{Shi:2016:ESPCN} based multi-scale scheme.
Specifically, from the input blurred image $B$ of size $W\times H$, we generate multi-scale input images as follows.
For the finest scale $S_3$, we use the input blurred image $B$.
The input image downsampled to a different scale is denoted by $B_i$, where $i$ is a scale index, i.e., $B_3 = B$, and $B_2$ is a downsampled version of $B$ of size $W/2 \times H/2$.

For $S_2$, we do not use $B_2$, but unshuffle $B_3$ to obtain four images of size $W/2 \times H/2$.
Then, we stack the unshuffled images along the channel direction to generate an input tensor $X_2$ for $S_2$.
As $B$ is an RGB image with three color channels, the size of $X_2$ is $W/2 \times H/2 \times 12$, so $X_2$ has the same spatial size as $B_2$ but still has the same amount of information as $B_3$.
Then, $X_2$ is fed into the feature extractor module ($E_2$ in \Fig{MSSNet-Test}) and processed through the stages at $S_2$.
Note that, despite $X_2$ having the same amount of information as $B_3$, the computation cost increase for $S_2$ is relatively small because we use features extracted from $X_2$ by the feature extractor module.
Moreover, thanks to $X_2$ having richer information than $B_2$, the sub-network at $S_2$ can produce a more accurate result.

For the coarsest scale $S_1$, we first downsample $B_3$ to obtain $B_2$. Then, we apply the same unshuffling process as for $S_2$ and obtain an input tensor $X_1$ for $S_1$.
Another possible choice is to directly unshuffle $B_3$ and obtain $X_1$ of $W/2 \times H/2 \times 48$, but we empirically found that this performs slightly worse.
While the pixel-shuffle-based multi-scale architecture can already enhance deblurring quality when trained with conventional loss terms as will be shown in \Sec{evaluation},
we propose a pixel-shuffle-based training strategy to minimize information loss and enhance deblurring quality in \Sec{training}.

\paragraph{Cross-Stage and Cross-Scale Feature Fusion}
MSSNet also adopts the cross-stage feature fusion scheme proposed in \cite{Zamir:2021:MPRNet}.
The cross-stage feature fusion scheme connects network modules in consecutive stages with additional connections (dotted pink lines in \Fig{MSSNet-Test}) to help information flow more effectively between stages.
\Fig{model_detail}(a) describes the cross-stage feature fusion scheme.
We refer the readers to~\cite{Zamir:2021:MPRNet} for more details on the cross-stage feature fusion scheme.
In addition, we also introduce cross-scale feature fusion (dotted green lines in \Fig{MSSNet-Test}) to facilitate more effective information flow between consecutive scales.
The cross-scale feature fusion scheme is described in \Fig{model_detail}(b).

\subsection{Training and Loss Functions}
\label{sec:training}

%%%%%%%%%%%%%%%%%%%%%%%%%%%%%%
\begin{figure*}[t]
    \begin{center}
    \includegraphics[width=0.8\linewidth]{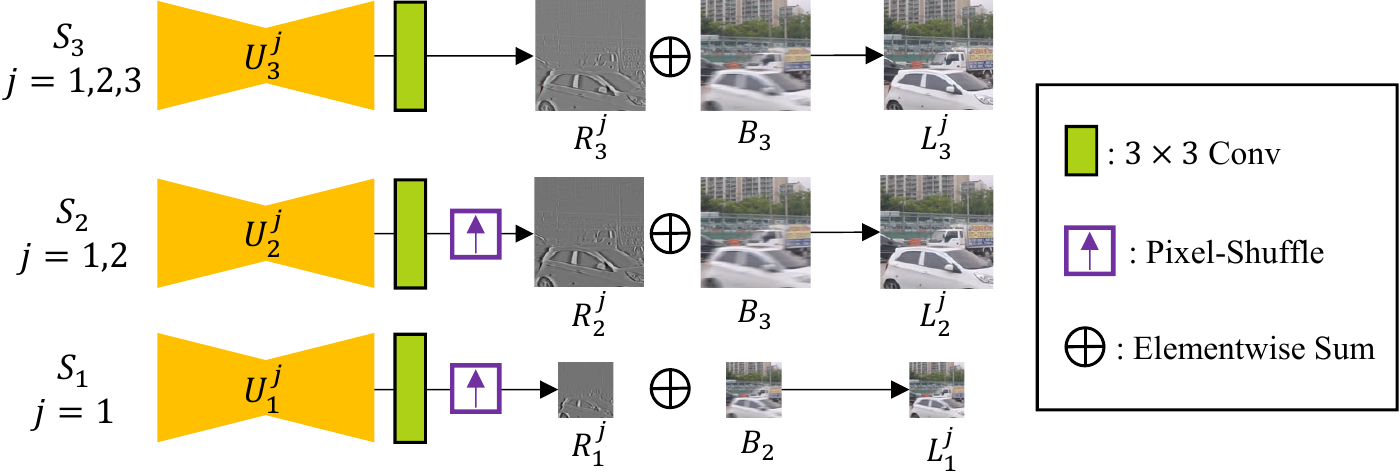}
    \end{center}
    \vspace{-9pt}
    \caption{Training of MSSNet. We train every stage to produce a residual image using auxiliary conv and pixel-shuffle layers.}
    \label{fig:training}
    \vspace{-7pt}
\end{figure*}
%%%%%%%%%%%%%%%%%%%%%%%%%%%%%

During training, we guide each stage of MSSNet to produce a deblurred image.
We attach auxiliary layers to every stage to produce a deblurred image, except for the last one in $S_3$ that already has such layers.
Specifically, for $S_3$, we attach an auxiliary conv layer at the end of $U_3^1$ and $U_3^2$ as shown in \Fig{training}.
The attached conv layers take features from the UNet modules and produce residual images $R_3^1$ and $R_3^2$.
Each residual image is then added to $B_3$ to produce deblurred results $L_3^1$ and $L_3^2$.
We also denote the final deblurred result $L$ by $L_3^3$.

For $S_1$ and $S_2$, we use a slightly different training strategy as the sub-networks at $S_1$ and $S_2$ take unshuffled images as input.
Specifically, at the end of each stage at $S_1$ and $S_2$, we attach a conv layer and a pixel-shuffle layer as shown in \Fig{training}.
The attached layers at the stages at $S_1$ and $S_2$ produce residual images of sizes $W/2 \times H/2$ and $W \times H$, respectively.
We denote the deblurred results from the auxiliary layers by $L_i^j$ where $i$ and $j$ are scale and stage indices, respectively.

We train MSSNet using two types of loss functions: a content loss $\mathcal{L}_{cont}$ and a frequency reconstruction loss $\mathcal{L}_{freq}$.
The content loss $\mathcal{L}_{cont}$ is defined as:
\begin{eqnarray}
    \mathcal{L}_{cont} &=& \frac{1}{N_1}\|L_1^1-L_{gt\downarrow}\|_1 + \sum\limits_{j=1}^2\frac{1}{N_2}\|L_2^j-L_{gt}\|_1 + \sum\limits_{j=1}^3\frac{1}{N_3}\|L_3^j-L_{gt}\|_1 ,
     \label{eq:L_cont} 
\end{eqnarray}
where $L_{gt}$ is the ground-truth blurred image, and $L_{gt\downarrow}$ is a downsampled version of $L_{gt}$.
$N_1$, $N_2$ and $N_3$ are normalization factors, which we set
$N_1 = W/2 \times H/2 \times 3$ and $N_2 = N_3 = W \times H \times 3$.
The frequency reconstruction loss was proposed in \cite{Cho:2021:MIMO} to restore high-frequency details from blurred image by minimizing the difference between blurred image and ground-truth in the frequency domain.
The frequency reconstruction loss is defined as:
\begin{eqnarray}
    \label{eq:L_freq}
    \mathcal{L}_{freq} &=& \frac{1}{N_1}\|\mathcal{F}(L_1^1)-\mathcal{F}(L_{gt\downarrow})\|_1
    + \sum\limits_{j=1}^2\frac{1}{N_2}\|\mathcal{F}(L_2^j)-\mathcal{F}(L_{gt})\|_1 \\
    && + \sum\limits_{j=1}^3\frac{1}{N_3}\|\mathcal{F}(L_3^j)-\mathcal{F}(L_{gt})\|_1 , \nonumber
\end{eqnarray}
where $\mathcal{F}$ is Fourier transform.
Finally, our final loss is %for training is defined as:
%\begin{equation}
$\mathcal{L}_{total} = \mathcal{L}_{cont} + \lambda \mathcal{L}_{freq}$
%\end{equation}
where $\lambda$ = 0.1.

\section{Experiments}
\label{sec:evaluation}

\subsection{Implementation Details}
\label{sec:implementation}

For evaluation, we trained MSSNet on the GoPro dataset~\cite{Nah:2017:DeepDeblur}.
For training, we randomly cropped $256\times256$ patches from blurry and sharp images.
Horizontal and vertical flips were randomly applied to cropped patches.
We trained our model for 3,000 epochs (396,000 iterations) with batch size 16.
We used the Adam optimizer~\cite{Kingma:2014:Adam} with cosine annealing~\cite{Loshchilov:2016:SGDR}.
We set the initial learning rate to $2\!\times\!10^{\texttt{-}4}$ and gradually decreased it to $1\!\times\!10^{\texttt{-}6}$.
To evaluate the performance of MSSNet on real-world blurred images,
we also use the RealBlur dataset~\cite{Rim:2020:RealBlur}.
For evaluation on the RealBlur test set,
We trained MSSNet using the GoPro~\cite{Nah:2017:DeepDeblur}, BSD-B~\cite{Rim:2020:RealBlur}, and RealBlur training sets following the RealBlur benchmark~\cite{Rim:2020:RealBlur}.
We trained the model for 100 epochs (397,400 iterations).
The other training details are the same as above.
The computation times of all models are measured on a PC with an NVIDA GeForce RTX 3090 GPU.

\subsection{Comparison with Previous Methods}
%%%%%%%%%%%%%%%%%%%%%%%%%%%%%%%%%%%%%%%
\begin{table}[t]
\renewcommand{\arraystretch}{1.1}
\begin{center}
\caption{Quantitative evaluation on the GoPro test dataset~\cite{Nah:2017:DeepDeblur}.
The models in \textcolor{blue}{blue} are coarse-to-fine approaches, while the models in \textcolor{purple}{red} are single-scale approaches. MIMO-UNet and its variants are based on a single UNet with multi-scale losses~\cite{Cho:2021:MIMO}.
The computation times of all the methods are measured in the same environment described in \Sec{implementation}.
The numbers of parameters, MACs, and computation times of RADN~\cite{Purohit:2020:RADN} and SAPHN~\cite{Suin:2020:SAPHN} are unavailable as their source codes are not publicly released yet.}
\label{tbl:evaluation_SOTA}
\setlength{\tabcolsep}{3pt} % Default value: 6pt
\scalebox{0.95}{
\begin{tabular}{l|c|c|c|c|c}
\toprule[1.2pt]
Models & PSNR (dB) & SSIM & Param (M) & MACs (G) & Time (s) \\ 
\hline
 \textcolor{blue}{DeepDeblur}~\cite{Nah:2017:DeepDeblur}& 29.08& 0.914& 11.72& 4729& 1.290  \\
 \textcolor{purple}{DMPHN}~\cite{Zhang:2019:DMPHN}& 30.25& 0.935& 7.23& 1100 & 0.137 \\
 \textcolor{blue}{SRN}~\cite{Tao:2018:SRN}& 30.26& 0.934& 8.06& 20134& 0.736 \\ 
 \textcolor{blue}{PSS-NSC}~\cite{Gao:2019:PSS-NSC}& 30.92& 0.942& 2.84& 3255& 0.316 \\
 \textcolor{purple}{MT-RNN}~\cite{Park:2020:MTRNN}& 31.15& 0.945& 2.6& 2315 & 0.323 \\
 \textcolor{purple}{SDNet4}~\cite{Zhang:2019:DMPHN}& 31.20& 0.945& 21.7& 3301 & 0.414 \\
 \textcolor{black}{MIMO-UNet}~\cite{Cho:2021:MIMO}& 31.73& 0.951& 6.8& 944 & 0.133 \\
 \textcolor{purple}{RADN}~\cite{Purohit:2020:RADN}& 31.76& 0.953& N/A& N/A& N/A\\
 \textcolor{purple}{SAPHN}~\cite{Suin:2020:SAPHN}& 32.02& 0.953& N/A& N/A& N/A\\
 \textcolor{blue}{MSSNet-small (Ours)} & 32.02& 0.953& 6.75& 634 & 0.104 \\
 \textcolor{black}{MIMO-UNet+}~\cite{Cho:2021:MIMO}& 32.45& 0.957& 16.1& 2171 & 0.290 \\
 \textcolor{purple}{MPRNet}~\cite{Zamir:2021:MPRNet}& 32.66& 0.959& 20.1& 10927 & 1.023 \\
 \textcolor{black}{MIMO-UNet++}~\cite{Cho:2021:MIMO}& 32.68& 0.959& 16.1& 8683 & 1.169 \\
 \textcolor{purple}{HINet}~\cite{Chen:2021:HINet}& 32.90& 0.960& 88.67& 2401 & 0.247 \\
%\hline\hline
\textcolor{blue}{MSSNet (Ours)} & 33.01& 0.961& 15.59& 2159 & 0.255 \\ 
\textcolor{blue}{MSSNet-large (Ours)} & 33.39& 0.964& 28.15& 4235 & 0.457 \\
\toprule[1.2pt]
\end{tabular}
}
\vspace{-9pt}
\end{center}
\end{table}
%%%%%%%%%%%%%%%%%%%%%%%%%%%%%%%%%%%%%%%

We compare MSSNet with previous state-of-the-art methods.
\Tbl{evaluation_SOTA} shows a quantitative comparison on the GoPro test set~\cite{Nah:2017:DeepDeblur}.
All the methods in the comparison were trained with the GoPro training set. 
Among the compared methods, DeepDeblur~\cite{Nah:2017:DeepDeblur}, SRN~\cite{Tao:2018:SRN} and PSS-NSC~\cite{Gao:2019:PSS-NSC} are coarse-to-fine approaches.
MIMO-UNet~\cite{Cho:2021:MIMO} is trained using multi-scale loss terms, but not a conventional coarse-to-fine approach as it is based on a single UNet architecture with an encoder that processes an image in a fine-to-coarse manner.
MIMO-UNet+ is a variant of MIMO-UNet with more parameters, and MIMO-UNet++ is MIMO-UNet+ with self-ensemble.
All the other methods are single-scale approaches.

As shown in \Tbl{evaluation_SOTA},
recent single-scale approaches tend to perform better than coarse-to-fine approaches except for MIMO-UNet~\cite{Cho:2021:MIMO} and its variants.
On the other hand, MSSNet clearly outperforms all the other methods in terms of PSNR and SSIM thanks to our remedies.
Specifically, MSSNet performs better than MIMO-UNet+ by more than 0.5dB with fewer parameters and fewer computations.
Compared to MIMO-UNet++, a self-ensemble version of MIMO-UNet+, MSSNet still outperforms by 0.33dB with a $4\times$ fewer computations.
Also, compared to HINet~\cite{Chen:2021:HINet}, MSSNet achieves 0.11dB higher PSNR with $5.7\times$ fewer parameters and fewer computations while slightly slower.

\begin{figure}[t]
    \begin{center}
    \includegraphics[width=0.95\linewidth]{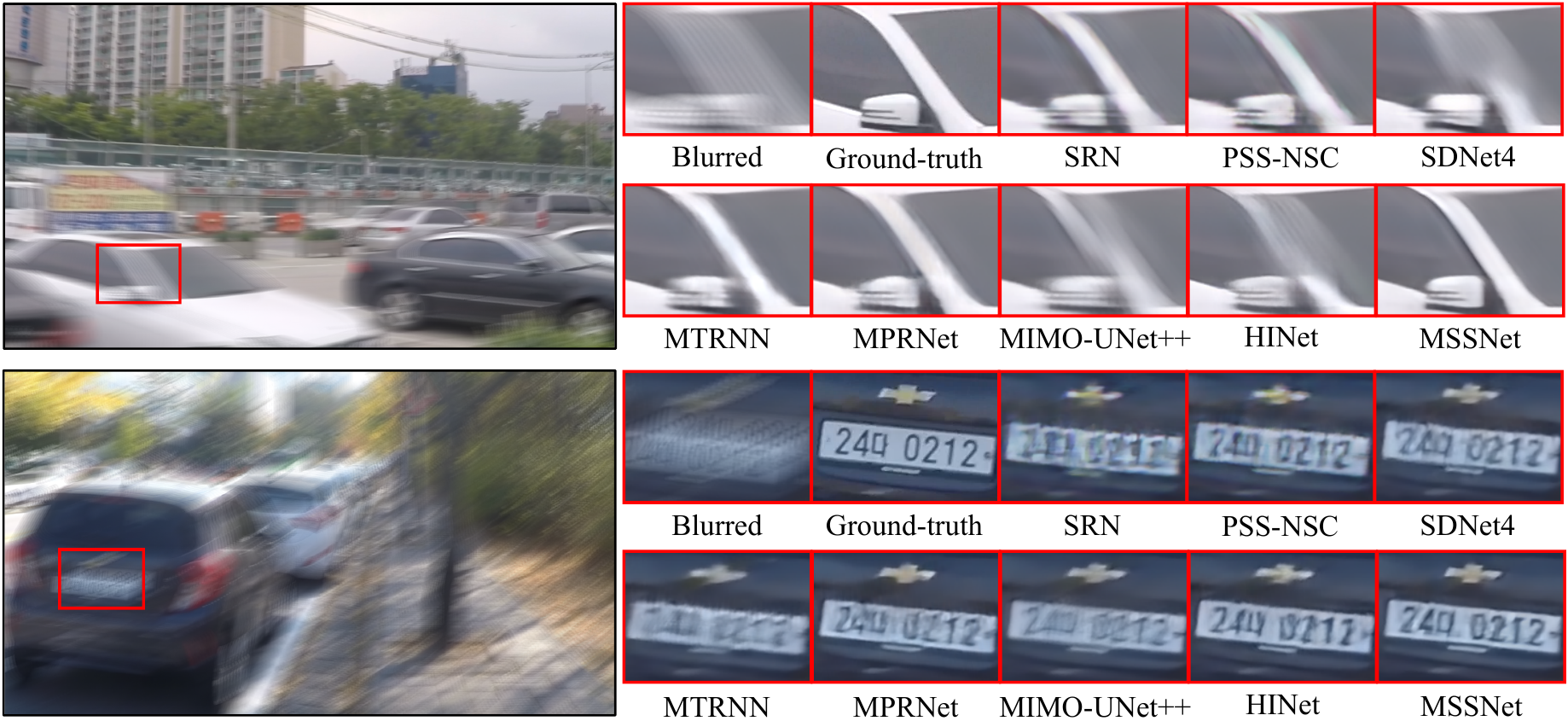}
    \end{center}
    \vspace{-0.3cm}
    \caption{Qualitative evaluation on the GoPro dataset~\cite{Nah:2017:DeepDeblur}.}
    \label{fig:GoPro_eval}
\vspace{-0.4cm}
\end{figure}

We also include two variants of MSSNet: MSSNet-small and MSSNet-large, in this evaluation.
Their detailed architectures are provided in the appendix.
Compared to MIMO-UNet and SRN, which have larger model sizes, MSSNet-small achieves a higher PSNR and SSIM with smaller computation time. 
While SAPHN~\cite{Suin:2020:SAPHN} achieves similar PSNR and SSIM values to those of MSSNet-small, ours performs much faster according to the computation time reported in their paper.
Specifically, the reported computation time of SAPHN measured on a Titan Xp GPU is 0.77 sec., while that of MSSNet-small on the same GPU is 0.19 sec.
MSSNet-large has about twice the parameters of MSSNet, which is still $3\times$ fewer than HINet, and its computation time is more than twice shorter than those of MIMO-UNet++ and MPRNet.
Nevertheless, it achieves 33.39 dB in PSNR, significantly exceeding all the other methods by a large margin.
\Fig{GoPro_eval} shows a qualitative comparison on the GoPro dataset~\cite{Nah:2017:DeepDeblur}. As shown in the figure, our results show clearly restored sharp details while those of the others have remaining blur.

%%%%%%%%%%%%%%%%%%%%%%%%%%%%%%%%%%%%%%%
\begin{table}[t]
\begin{center}
\caption{Quantitative evaluation on RealBlur~\cite{Rim:2020:RealBlur}. The models in the upper part of the table are trained on the GoPro dataset~\cite{Nah:2017:DeepDeblur} and tested on the RealBlur test sets. The models in the lower part are trained and tested on each of the RealBlur-R and RealBlur-J datasets.
MIMO-UNet++~\cite{Cho:2021:MIMO} provides only a model trained on the RealBlur-J dataset.
}
\label{tbl:evaluation_realblur}
\setlength{\tabcolsep}{6pt} % Default value: 6pt
\scalebox{1}{
\begin{tabular}{l|c|c|c|c}
\toprule[1.2pt]
Models & \multicolumn{2}{c|}{RealBlur-R}& \multicolumn{2}{c}{RealBlur-J} \\
& \multicolumn{1}{c|}{PSNR} & \multicolumn{1}{c|}{SSIM} & \multicolumn{1}{c|}{PSNR} & \multicolumn{1}{c}{SSIM} \\ 
\toprule[1.2pt]
Hu \etal~\cite{Hu:2014:deblurring}& 33.67& 0.916& 26.41& 0.803\\ 
DeepDeblur~\cite{Nah:2017:DeepDeblur}& 32.51& 0.841& 27.87& 0.827 \\
DeblurGAN~\cite{Kupyn:2018:DeblurGan}& 33.79& 0.903& 27.97& 0.834 \\
Pan \etal~\cite{Pan:2016:blind}& 34.01& 0.916& 27.22& 0.790\\
Xu \etal~\cite{Xu:2013:unnatural}& 34.46& 0.937& 27.14& 0.830\\
DeblurGAN-v2~\cite{Kupyn:2019:deblurgan}& 35.26& 0.944& 28.70& 0.866\\
Zhang \etal~\cite{Zhang:2018:dynamic}& 35.48& 0.947& 27.80& 0.847\\
SRN~\cite{Tao:2018:SRN}& 35.66& 0.947& 28.56& 0.867\\
SDNet4~\cite{Zhang:2019:DMPHN}& 35.70& 0.948& 28.42& 0.860\\
MPRNet~\cite{Zamir:2021:MPRNet}&35.99 &0.952 &28.70 &0.873 \\
\textbf{MSSNet (Ours)}&35.93 &0.953 & 28.79&0.879 \\ 
\hline\hline
DeblurGAN-v2~\cite{Kupyn:2019:deblurgan}&36.44 &0.935 &29.69 &0.870 \\
SRN~\cite{Tao:2018:SRN}& 38.65& 0.965& 31.38& 0.909\\
MPRNet~\cite{Zamir:2021:MPRNet}& 39.31& 0.972& 31.76&0.922 \\
MIMO-UNet++~\cite{Cho:2021:MIMO} & N/A & N/A & 32.05&0.921 \\
\textbf{MSSNet (Ours)}& 39.76& 0.972& 32.10&0.928 \\
\toprule[1.2pt]
\end{tabular}
}
\end{center}
    \vspace{-0.4cm}
\end{table}
%%%%%%%%%%%%%%%%%%%%%%%%%%%%%%%%%%%%%%%

%%%%%%%%%%%%%%%%%%%%%%%%%%%%%%%%%%%%%%%%
%%%%%%%%%%%%%%%%%%%%%%%%%%%%%%%%%%%%%%%%%

%%%%%%%%%%%%%%%%%%%%%%%%%%%%%%%%%%%%%%%%%%
\begin{figure}[t]
    \begin{center}
    \includegraphics[width=0.95\linewidth]{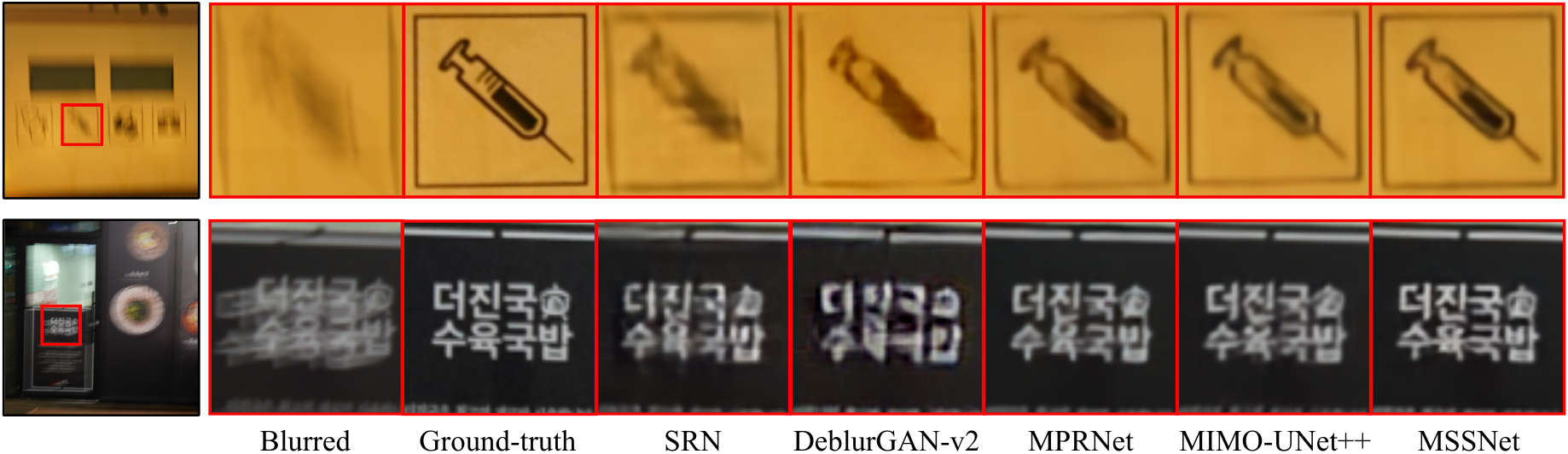}
    \end{center}
    \vspace{-0.3cm}
    \caption{Qualitative evaluation on the ReaBlur-J dataset~\cite{Rim:2020:RealBlur}.}
    \label{fig:RealBlurJ}
\vspace{-0.2cm}
\end{figure}
%%%%%%%%%%%%%%%%%%%%%%%%%%%%%%%%%%%%%%%%%%%

We also study the generalization ability and performance of MSSNet on real-world blurred images.
\Tbl{evaluation_realblur} shows a quantitative evaluation on the RealBlur dataset~\cite{Rim:2020:RealBlur}, which consists of real-world blurred images.
The methods in the upper section in the table are trained on the GoPro dataset~\cite{Nah:2017:DeepDeblur},
while those in the lower section are trained on the RealBlur-R and RealBlur-J datasets.
Among the methods trained on the GoPro datasets, MSSNet achieves the highest SSIM for the RealBlur-R test set, and the highest PSNR and SSIM for the RealBlur-J test set.
Also, among the methods trained on the RealBlur datasets, MSSNet achieves the highest PSNR and SSIM.
\Fig{RealBlurJ} shows a qualitative comparison on the RealBlur-J dataset~\cite{Rim:2020:RealBlur}.
In all the examples, the results of the other methods show either remaining blur and incorrectly restored details.
On the other hand, our results show better restored details.
Additional qualitative examples are provided in the appendix.

\subsection{Ablation Study and Analysis}
We validate the effectiveness of the coarse-to-fine approach, and then analyze the effect of each technical component in our model.
For analysis, we test several variants of MSSNet. All the models in the analysis are trained and tested on the GoPro training and test sets~\cite{Nah:2017:DeepDeblur}, respectively.
For ease of analysis, all the variants of MSSNet in the ablation studies use neither the pixel-shuffling scheme nor the cross-stage and cross-scale feature fusion scheme if not otherwise noted.

%%%%%%%%%%%%%%%%%%%%%%%%%%%%%%%%%
\begin{table}[t]
\begin{center}
\setlength\tabcolsep{3.3pt}
\caption{Performance comparison among a single-scale architecture with four stages and our multi-scale architectures.
MSSNet-Single is a single-scale architecture, while MSSNet-Multi and MSSNet-Multi-Small are multi-scale architectures.
`Initial' and `Final' are the initial and final results of each architecture, respectively.
}
\label{tbl:ab_mstage_mscale1}
\begin{tabular}{l|c|c|c}
\toprule
& MSSNet-Single & MSSNet-Multi-Small & MSSNet-Multi                          \\
\midrule[0.4pt]
PSNR (Initial / Final) & 29.11 / 31.59 & 29.51 / 31.58 &  30.09 / 31.75 \\
\midrule[0.1pt]
Params (M) / MACs (G) & 4.39 / 660.69 & 4.38 / 574.82 & 6.61 / 621.60 \\
\bottomrule
\end{tabular}
\vspace{-0.4cm}
\end{center}
\end{table}
%%%%%%%%%%%%%%%%%%%%%%%%%%%%%%%%%%%%%%%%

\paragraph{Coarse-to-Fine \emph{vs} Single-Scale}
As discussed in \Sec{intro}, the coarse-to-fine approach can quickly estimate a high-quality initial solution using coarse scales.
Specifically, compared to performing a single stage of deblurring at the original scale,
performing multiple stages at a coarse scale can be computationally more efficient.
Moreover, thanks to the small blur size at a coarse scale, it can estimate a more accurate result, which serves as an initial solution for a finer scale, which leads to a final deblurring result of higher quality.

To verify this, in \Tbl{ab_mstage_mscale1}, we compare three variants of MSSNet.
MSSNet-Single is a single-scale model with four stages at the original scale.
MSSNet-Multi and MSSNet-Multi-Small are multi-scale models with the same number of scales and stages as MSSNet.
MSSNet-Single and MSSNet-Multi has the same number of parameters for each stage.
On the other hand, MSSNet-Multi-Small has fewer parameters for each stage at $S_1$ and $S_2$ so that its total number of parameters is similar to that of MSSNet-Single. Its architecture details are in the appendix.
The multi-scale models use our pixel-shuffle-based approach, but none of the models use the cross-stage and cross-scale feature fusion schemes.
While the multi-scale models have six stages in total, three of them are at coarser scales.
As a result, both multi-scale model require smaller amounts of computation than MSSNet-Single as shown in the table.

In \Tbl{ab_mstage_mscale1}, `Initial' and `Final' indicates the initial and final results of the single-scale and multi-scale models.
An initial solution of the single-scale model indicates a deblurring result of the first stage obtained using an auxiliary conv layer,
while an initial solution of the multi-scale models indicates a deblurring result of the last stage at $S_2$ obtained using auxiliary conv and pixel-shuffle layers.
We compare these as they serve as initial solutions for the last three stages.
As shown in the table, despite its smaller computation cost, MSSNet-Multi produces higher-quality initial and final deblurring results.
Also, although MSSNet-Multi-Small has a similar number of parameters and a much smaller computation cost, it still achieves a similar PSNR for the final result to that of MSSNet-Single.
This proves the advantage of the coarse-to-fine approach against the single-scale approach.
%%%%%%%%%%%%%%%%%%%%%%%%%%%%%%%%%%%%%%%%%%%%%%%%%%%%%%%%%%%%%%%%

%%%%%%%%%%%%%%%%%%%%%%%%%%%%%%%%%%
\setlength{\tabcolsep}{4pt}
\begin{table}[t]
\renewcommand{\arraystretch}{1.1}
\begin{center}
\caption{Ablation study on the stage configuration using variants of DeepDeblur~\cite{Nah:2017:DeepDeblur}.}
%\vspace{-0.2cm}
\label{tbl:stage_confi_deepdeblur}
\begin{tabular}{c|c|c|c|c|c|c|c}
\toprule[1.2pt]
       & \multicolumn{3}{c|}{\# ResBlocks} &  &  &  &  \\ 
Models & $S_1$ & $S_2$ & $S_3$ & PSNR & SSIM & Params (M) & MACs (G) \\ \hline
D444 & 4 & 4 & 4 & 27.07 & 0.8269 & 2.5 &  1009.3 \\
D444L & 4 & 4 & 4 & 27.26 & 0.8315 & 3.42 & 1382.3 \\
D246 & 2 & 4 & 6 & 27.38 & 0.8324 & 2.5 & 1363.5  \\
\toprule[1.2pt]
\end{tabular}
\end{center}
\vspace{-0.4cm}
\end{table}
\setlength{\tabcolsep}{1.4pt}
%%%%%%%%%%%%%%%%%%%%%%%%%%%%%%%%%%%

%%%%%%%%%%%%%%%%%%%%%%%%%%%%%%%%%%
\setlength{\tabcolsep}{4pt}
\begin{table}[t]
\renewcommand{\arraystretch}{1.1}
\begin{center}
\caption{Ablation study on the stage configuration using variants of MSSNet.}
%\vspace{-0.2cm}
\label{tbl:stage_confi}
\begin{tabular}{c|c|c|c|c|c|c|c}
\toprule[1.2pt]
      & \multicolumn{3}{c|}{\# Stages} &  &  &  &   \\
Models & $S_1$ & $S_2$ & $S_3$ & PSNR & SSIM & Params (M) & MACs (G) \\ \hline
M123 & 1 & 2 & 3 & 29.58 & 0.925 & 1.18 & 521.33  \\
M552 & 5 & 5 & 2 & 29.27 & 0.920 & 1.18 & 521.33 \\
\toprule[1.2pt]
\end{tabular}
\end{center}
\vspace{-0.4cm}
\end{table}
\setlength{\tabcolsep}{1.4pt}
%%%%%%%%%%%%%%%%%%%%%%%%%%%%%%%%%%%

\paragraph{Stage Configuration Reflecting Blur Scales}
Our first remedy that we adopt into our MSSNet is the stage configuration reflecting blur scales.
To verify its effect as a common rule, we conduct two ablation studies using DeepDeblur~\cite{Nah:2017:DeepDeblur} and MSSNet.

\Tbl{stage_confi_deepdeblur} compares three variants of DeepDeblur~\cite{Nah:2017:DeepDeblur}.
D444 and D444L have four residual blocks at each scale, while D246 adopts our stage configuration scheme and has two, four and six residual blocks at $S_1$, $S_2$ and $S_3$, respectively.
To match the computation cost of D246, we also prepare D444L, which has more channels at each residual block.
The table shows that D246 outperforms both of the others in terms of PSNR and SSIM,
especially, despite its fewer parameters and a smaller computation cost than those of D444L.

In the second experiment, we compare two variants of MSSNet in \Tbl{stage_confi}.
The variants have different numbers of stages at different scales as informed in the table, but share the same network architecture for the UNet modules.
The deblurring performance is not only affected by the number of stages, but also by the computation amount and the number of parameters.
To isolate the impact of the stage configuration on the deblurring performance from other factors,
each of the tested models in this experiment shares the network weights across different stages.
In \Tbl{stage_confi}, M123 has the same stage configuration as MSSNet.
M552 has fewer stages at $S_3$ but more stages at coarse scales so it requires the same amount of computation.
The table shows that M123 clearly outperforms M552, validating our argument on the stage configuration.
Additional experiments with different settings, e.g., models without parameter sharing, are provided in the appendix.
%%%%%%%%%%%%%%%%%%%%%%%%%%%%%%%%%%%%%%%%%%%%%%%%%%%%%%%%%%%%%%

%%%%%%%%%%%%%%%%%%%%%
\setlength{\tabcolsep}{4pt}
\begin{table}[t]
\begin{center}
\caption{Ablation study on the scale information propagation.}
\label{tbl:sol_feature}
\begin{tabular}{l|c|c|c|c}
\toprule[1.2pt]
Model & PSNR & SSIM &Params (M) & MACs (G)\\ \hline
MSS(Image,Concat) &31.42 &0.947 & 6.59 & 613.1 \\
MSS(Feature,Skip) &31.52 &0.948 & 6.59 & 621.8 \\
MSS(Feature,Concat) &31.54 &0.949 & 6.61 & 621.1 \\
\toprule[1.2pt]
\end{tabular}
\end{center}
\vspace{-0.4cm}
\end{table}
\setlength{\tabcolsep}{1.4pt}
%%%%%%%%%%%%%%%%%%%%%

\paragraph{Inter-Scale Feature Propagation}
In the next ablation study, we verify the effect of our inter-scale feature propagation scheme.
In this study, we also investigate how to fuse the solution from a coarse scale with the input to the finer scale.
To this end, we compare three variants of MSSNet: MSS(Image,Concat), MSS(Feature,Skip) and MSS(Feature,Concat).
MSS(Image,Concat) has auxiliary conv layers at the end of $S_1$ and $S_2$ to convert features to residual images. The residual images are added to the input blurred images of the corresponding sizes to produce deblurred results. The deblurred results are then upsampled and concatenated to the blurred images at the next scales.
This model corresponds to the previous coarse-to-fine approaches that transfer pixel values from coarse to fine scales.
MSS(Feature,Skip) transfers features from coarse to fine scales as done in MSSNet. However, features from coarse scales are not concatenated but added to the features of the blurred images at the next scales.
As the sub-networks estimate residual features, adding them to the features of blurred images will produce initial deblurred features at finer scales.
MSS(Feature,Concat) uses our inter-scale feature propagation scheme that concatenates features from coarse scales to the features of the blurred images at the next scales.

\Tbl{sol_feature} compares the performance of the variants.
The results confirm that using features instead of pixel values clearly improves the deblurring quality as features provide richer information.
The table also shows that MSS(Feature,Concat) performs slightly better than MSS(Feature,Skip), although it requires slightly more parameters, validating our approach.
%%%%%%%%%%%%%%%%%%%%%%%%%%%%%%%%%%%%%%%%%%%%%%%%%%%%%%%%%%%%

%%%%%%%%%%%%%%%%%%%%%%%
\setlength{\tabcolsep}{4pt}
\begin{table}[t]
\begin{center}
\caption{Ablation study on the pixel-shuffle-based multi-scale approach. PUS: pixel-unshuffle. PS: pixel-shuffle.}
\label{tbl:pixel_shffle}
\begin{tabular}{cc|c|c|c|c}
\toprule[1.2pt]
PUS& PS & PSNR & SSIM & Params (M) & MACs (G) \\ \hline
& &31.54 &0.949 &6.61 & 621.1 \\
\checkmark & &31.67 &0.950 &6.61 & 621.6 \\
\checkmark & \checkmark &31.75 &0.951 &6.61 & 621.6\\ \toprule[1.2pt] 
\end{tabular}
\end{center}
\vspace{-0.4cm}
\end{table}
\setlength{\tabcolsep}{1.4pt}
%%%%%%%%%%%%%%%%%%%%%%%%

\paragraph{Pixel-Shuffle-Based Multi-Scale Scheme}
We then verify the effect of our pixel-shuffle-based multi-scale scheme.
As discussed in \Sec{MSSNet}, our pixel-shuffle-based multi-scale scheme consists of pixel-unshuffle layers that generate input tensors, and auxiliary pixel-shuffle layers used only in the training phase.
To verify the effect of each component, we compare the performance of three variants of MSSNet: 1) without both pixel-unshuffle and shuffle layers, 2) with only the pixel-unshuffle layers, and 3) with both layers in \Tbl{pixel_shffle}.
The first model takes downsampled images as input as done in previous coarse-to-fine approaches, and its sub-networks at $S_1$ and $S_2$ are trained to produce intermediate results of the corresponding sizes.
The second model takes tensors generated by pixel-unshuffling layers as input, but its sub-networks in $S_1$ and $S_2$ are trained in the same manner as the first model.
The third model corresponds to our approach.

As \Tbl{pixel_shffle} shows, introducing the pixel-unshuffling and shuffling layers introduces a negligible increase in the number of parameters.
On the other hand, the pixel-unshuffling layers clearly improve the deblurring quality as they provide richer information than downsampling.
Also, the auxiliary pixel-shuffling layers further improve the deblurring quality as they enable higher-quality supervision.

\section{Conclusion}
In this work, we analyzed the defects of previous deep learning-based coarse-to-fine approaches to single image deblurring.
Based on our analysis, we proposed MSSNet, a novel coarse-to-fine approach with our remedies to the defects.
MSSNet adopts stage configuration reflecting blur scales, inter-scale feature propagation, and pixel-shuffle-based multi-scale network architecture.
The experiment results prove the effectiveness of our novel technical components and show that our method is superior compared to the previous state-of-the art methods in regard to the accuracy, computation time, and network size.

\paragraph{Limitations and Future Work}
While MSSNet achieves the state-of-the-art performance, it still fails on many real-world blurred images especially with large blur as other methods.
Extending MSSNet for handling large blur can be an interesting future work.
Also, improving the computational efficiency and reducing the model size to deploy deblurring on mobile devices can be another interesting direction.
As a future work, we also plan to examine the performance of MSSNet on other types of image degradation.

\clearpage

\setcounter{figure}{0}
\setcounter{table}{0}

\renewcommand\thefigure{A\arabic{figure}}
\renewcommand\thetable{A\arabic{table}}

\appendix
\section{Appendix}
%****************************************
\subsection{Network Architectures}
\vspace{-0.1cm}
The detailed architectures of MSSNet, MSSNet-small and MSSNet-large are shown in \Tbls{Res}, \ref{tbl:UNet} and \ref{tbl:MSSNet}.
\Tbl{symbols} defines symbols used in the tables.
In this section, CSFF represents both cross-stage~\cite{Zamir:2021:MPRNet} and cross-scale feature fusion.
The architecture of UNet is based on that of MPRNet~\cite{Zamir:2021:MPRNet} without channel attention.

\paragraph{Our Models}
For MSSNet, the channel sizes $x$, $y$ and $z$ of UNet are set to 54, 96 and 138, respectively.
For MSSNet-small, they are set to 20, 60 and 100, respectively.
For MSSNet-large, they are set to 80, 130 and 180, respectively.

\paragraph{Models in the ablation study (Table 3. in the paper)}
For the models in the ablation study, CSFF is not used (i.e., set to false) in all of the scales in \Tbl{MSSNet}.  
For MSSNet-Multi, the channel sizes $x$, $y$ and $z$ are set to 20, 60 and 100, respectively in all of the UNet modules.
In MSSNet-Multi-Small, the channel sizes of UNet are different according to the scales.
Specifically, $x$, $y$ and $z$ are set to 20, 36 and 52, respectively, in $U_1^1$, $U_2^1$ and $U_2^2$,
and set to 20, 60 and 100, respectively, in $U_3^1$, $U_3^2$ and $U_3^3$.
%%%%%%%%%%%%%%%%%%%%%%%%%%%%%%%%%%%%%%%%%%%%%
\setlength{\tabcolsep}{4pt}
\begin{table}[t]
\renewcommand{\arraystretch}{1.1}
\begin{center}
\caption{Descriptions of the symbols in \Tbls{Res}, \ref{tbl:UNet} and \ref{tbl:MSSNet}.}
\label{tbl:symbols}
\scalebox{0.95}{
\begin{tabular}{c| c}
\toprule[1.2pt]
Symbols & Description \\
\hline
\hline
\texttt{\blue{type}} & Layer type \\
\texttt{\blue{input}} & Input feature name \\
\texttt{\blue{k}} & Filter size of a conv layer ($k\times k$) \\
\texttt{\blue{c}} & Output channel size of a layer \\
\texttt{\blue{s}} & Stride \\
\texttt{\blue{p}} & Padding size \\
\texttt{\blue{r}} & Upsampling or downsampling ratio \\
\multirow{2}{*}{\texttt{\blue{CSFF}}} & Whether to use cross-stage or scale  \\
                                      & feature fusion, or not \\
\texttt{\blue{output}} & Output feature name \\
\hline
\textrm{\violet{conv}} & Convolution layer (bias = False) \\
\textrm{\violet{PRelu}} & Parametric ReLU layer \\
\textrm{\violet{sum}} & Element-wise summation \\
\textrm{\violet{bi-down}} & Bilinear downsampling \\
\textrm{\violet{bi-up}} & Bilinear upsampling \\
\textrm{\violet{concat}} & Concatenation \\
\textrm{\violet{unshuffle}} & Pixel-unshuffle \\
\textrm{\violet{shuffle}} & Pixel-shuffle \\
\textrm{\violet{$U_i^j$}} & UNet module ($j$-th stage, $i$-th scale) \\ 
\textrm{\violet{Res}} & Residual block \\
\toprule[1.2pt]
\end{tabular}
}
\end{center}
\end{table}
%%%%%%%%%%%%%%%%%%%%%%%%%%%%%%%%%%%%%%%%%%%%%%

%%%%%%%%%%%%%%%%%%%%%%%%%%%%%%%%%%%%%%%%%%%%%%
%Resblock
\setlength{\tabcolsep}{4pt}
\begin{table}[t]
\renewcommand{\arraystretch}{1.1}
\begin{center}
\caption{Detailed architecture of a residual block (\textrm{\violet{Res}}). The channel size $n$ is a variable that is set differently at different locations in the MSSNet. Refer to \Tbls{UNet} and \ref{tbl:MSSNet}.
}
\label{tbl:Res}
\scalebox{0.95}{
\setlength{\tabcolsep}{3pt} % Default value: 6pt
\begin{tabular}{c| c| c| c| c| c| c}
\toprule[1.2pt]
\texttt{\blue{type}} & \texttt{\blue{input}} & \texttt{\blue{k}} & \texttt{\blue{c}} & \texttt{\blue{s}} & \texttt{\blue{p}} & \texttt{\blue{output}} \\
\hline
\hline
\textrm{\violet{conv}} & \textit{feat} & 3 & $n$ & 1 & 1 & $conv_1$ \\
\textrm{\violet{PRelu}} & $conv_1$ & - & $n$ & - & - & \textit{act} \\
\textrm{\violet{conv}} & \textit{act} & 3 & $n$ & 1 & 1 & $conv_2$ \\
\textrm{\violet{sum}} & \textit{feat}, $conv_2$& - & $n$ & - & - &  \textit{out} = \textit{feat} + $conv_2$\\
\toprule[1.2pt]
\end{tabular}
}
\end{center}
\end{table}
%%%%%%%%%%%%%%%%%%%%%%%%%%%%%%%%%%%%%%%%%%%%%%

%%%%%%%%%%%%%%%%%%%%%%%%%%%%%%%%%%%%%%%%%%%%%%%%%%%%
%UNet
\begin{table}[t]
\renewcommand{\arraystretch}{1.2}
\begin{center}
\caption{Detailed architecture of \textbf{UNet} in each stage at each scale (\textrm{\violet{$U_{1}^{1}$}}, \textrm{\violet{$U_{2}^{1}$}}, \textrm{\violet{$U_{2}^{2}$}}, \textrm{\violet{$U_{3}^{1}$}}, \textrm{\violet{$U_{3}^{2}$}}, and \textrm{\violet{$U_{3}^{3}$}}).
A UNet module takes an input feature tensor denoted by $feat$.
The layers in the \textbf{CSFF} sub-block are used only when \texttt{\blue{CSFF}} in \Tbl{MSSNet} is true.
In such a case, a UNet module utilizes additional features from the previous stage or scale, $[{p~el}_1, {p~el}_2, {p~el}_3]$ and $[{p~dl}_1, {p~dl}_2, {p~dl}_3]$
where ${p~el}_i$ and ${p~dl}_i$ correspond to ${el}_i$ and ${dl}_i$  in the previous stage or scale.
For \violet{$U_{1}^{1}$}, \violet{$U_{2}^{1}$} and \violet{$U_{3}^{1}$}, ${feat}$ is set to ${feat}_1$, ${fusion}_{12}$ and ${fusion}_{23}$ in \Tbl{MSSNet}, respectively.
Except for them, ${feat}$ is set to ${p~dl}_1$.
}
\label{tbl:UNet}
\begin{adjustbox}{max width=\textwidth}
\begin{tabular}{c|c|c|c|c|c|c|c|c}
\toprule[1.2pt]
Network sub-blocks & \texttt{\blue{type}} & \texttt{\blue{input}} & \texttt{\blue{k}} & \texttt{\blue{c}} & \texttt{\blue{s}} & \texttt{\blue{p}} & \texttt{\blue{r}} & \texttt{\blue{output}} \\
\hline
\hline
\multirow{2}{*}{\textbf{Encoder level 1}} & \textrm{\violet{Res}} & \textit{feat} & 3 & $x$ & 1 & 1 & - & $res_1$ \\ 
& \textrm{\violet{Res}} & $res_1$ & 3 & $x$ & 1 & 1 & - & ${el}_1$ \\
\hline
\multirow{3}{*}{\textbf{CSFF}} 
% & {$bi~upsample^*$}
% &${p~el}_1$, ${p~dl}_1$ & - & a & - & - & 2 & ${p~el}_1$, ${p~dl}_1$ \\
&\textrm{\violet{conv}}
& ${p~el}_1$ & 1 & $x$ & 1 & 0 & - & ${pc~el}_1$ \\
&\textrm{\violet{conv}}
& ${p~dl}_1$ & 1 & $x$ & 1 & 0 & - & ${pc~dl}_1$ \\
& \textrm{\violet{sum}} & ${el}_1$, ${pc~el}_1$, ${pc~dl}_1$ & - & $x$ & - & - & - &  ${el}_1$ = ${el}_1$ + ${pc~el}_1$ + ${pc~dl}_1$\\
\hline
\multirow{2}{*}{\textbf{Down}} & \textrm{\violet{bi-down}} & ${el}_1$ & - & $x$ & - & - & 0.5 & $down~{el}_1$ \\
& \textrm{\violet{conv}} & $down~{el}_1$ & 1 & $y$ & 1 & 0 & - & $down~{el}_1$ \\
\hline
\multirow{2}{*}{\textbf{Encoder level 2}} & \textrm{\violet{Res}} & $down~{el}_1$ & 3 & $y$ & 1 & 1 & - & $res_2$ \\ 
&\textrm{\violet{Res}}& $res_2$ & 3 & $y$ & 1 & 1 & - & ${el}_2$ \\
\hline
\multirow{3}{*}{\textbf{CSFF}} 
% & {$bi~upsample^*$}
% &${p~el}_2$, ${p~dl}_2$ & - & a + b & - & - & 2 & ${p~el}_2$, ${p~dl}_2$ \\ 
&\textrm{\violet{conv}}
& ${p~el}_2$ & 1 & $y$ & 1 & 0 & - & ${pc~el}_2$ \\
&\textrm{\violet{conv}}
& ${p~dl}_2$ & 1 & $y$ & 1 & 0 & - & ${pc~dl}_2$ \\
& \textrm{\violet{sum}} & ${el}_2$, ${pc~el}_2$, ${pc~dl}_2$ & - & $y$ & - & - & - &  ${el}_2$ = ${el}_2$ + ${pc~el}_2$ + ${pc~dl}_2$\\
\hline
\multirow{2}{*}{\textbf{Down}} & \textrm{\violet{bi-down}} & ${el}_2$ & - & $y$ & - & - & 0.5 & $down~{el}_2$ \\
& \textrm{\violet{conv}} & $down~{el}_2$ & 1 & $z$ & 1 & 0 & - & $down~{el}_2$ \\
\hline
\multirow{2}{*}{\textbf{Encoder level 3}} & \textrm{\violet{Res}} & $down~{el}_2$ & 3 & $z$ & 1 & 1 & - & $res_3$ \\ 
&\textrm{\violet{Res}}& $res_3$ & 3 & $z$ & 1 & 1 & - & ${el}_3$ \\
\hline
\multirow{3}{*}{\textbf{CSFF}} 
&\textrm{\violet{conv}}
& ${p~el}_3$ & 1 & $z$ & 1 & 0 & - & ${pc~el}_3$ \\
&\textrm{\violet{conv}}
& ${p~dl}_3$ & 1 & $z$ & 1 & 0 & - & ${pc~dl}_3$ \\
& \textrm{\violet{sum}} & ${el}_3$, ${pc~el}_3$, ${pc~dl}_3$ & - & $z$ & - & - & - &  ${el}_3$ = ${el}_3$ + ${pc~el}_3$ + ${pc~dl}_3$\\
\hline
\multirow{2}{*}{\textbf{Decoder level 3}} & \textrm{\violet{Res}} & ${el}_3$ & 3 & $z$ & 1 & 1 & - & $res_4$ \\ 
&\textrm{\violet{Res}}& $res_4$ & 3 & $z$ & 1 & 1 & - & ${dl}_3$ \\
\hline
\multirow{2}{*}{\textbf{Up}} & \textrm{\violet{bi-up}} & ${dl}_3$ & - & $z$ & - & - & 2 & $up~{dl}_3$ \\
& \textrm{\violet{conv}} & $up~{dl}_3$ & 1 & $y$ & 1 & 0 & - & $up~{dl}_3$ \\
\hline
\multirow{2}{*}{\textbf{Skip}} & \textrm{\violet{Res}} & ${el}_2$ & 3 & $y$  & 1 & 1 & - & ${skip}~{el_2}$ \\
& \textrm{\violet{sum}} & ${skip}~{el_2}$, $up~{dl}_3$ & - & $y$ & - & - & - & $ up~{dl}_3 = {skip}~{el_2}$ + $up~{dl}_3$ \\
\hline
\multirow{2}{*}{\textbf{Decoder level 2}} & \textrm{\violet{Res}} & ${up~dl}_3$ & 3 & $y$ & 1 & 1 & - & $res_5$ \\ 
&\textrm{\violet{Res}}& $res_5$ & 3 & $y$ & 1 & 1 & - & ${dl}_2$ \\
\hline
\multirow{2}{*}{\textbf{Up}} & \textrm{\violet{bi-up}} & ${dl}_2$ & - & $y$ & - & - & 2 & $up~{dl}_2$ \\
& \textrm{\violet{conv}} & $up~{dl}_2$ & 1 & $x$ & 1 & 0 & - & $up~{dl}_2$ \\
\hline
\multirow{2}{*}{\textbf{Skip}} & \textrm{\violet{Res}} & ${el}_1$ & 3 & $x$ & 1 & 1 & - & ${skip}~{el_1}$ \\
& \textrm{\violet{sum}} & ${skip}~{el_1}$, $up~{dl}_2$ & - & $x$ & - & - & - & $ up~{dl}_2 = {skip}~{el_1}$ + $up~{dl}_2$ \\
\hline
\multirow{2}{*}{\textbf{Decoder level 1}} & \textrm{\violet{Res}} & $up~{dl}_2$ & 3 & $x$ & 1 & 1 & - & $res_6$ \\ 
&\textrm{\violet{Res}}& $res_6$ & 3 & $x$ & 1 & 1 & - & ${dl}_1$ \\
\toprule[1.2pt]
\end{tabular}
\end{adjustbox}
\end{center}
\end{table}

%%%%%%%%%%%%%%%%%%%%%%%%%%%%%%%%%%%%%%%%%%%%%%%%%%%%%%%%%%%%%%%%%%%%%%

% MSSNet
\begin{table}[t]
\renewcommand{\arraystretch}{1.3}
\begin{center}
\caption{Detailed architecture of MSSNet.
\textbf{Auxiliary} layers are only used for training.
The sub-blocks $E_1$, $E_2$ and $E_3$ correspond to $E_1$, $E_2$ and $E_3$ in Fig.~2, respectively.
While \violet{$U_1^1$} takes only one input tensor ${feat}_1$, the other UNet modules take three input tensors as they use CSFF.
The first input of each UNet module corresponds to ${feat}$ in \Tbl{UNet}, while the second and third input values correspond to $\{{p~el}_i\}$ and $\{{p~dl}_i\}$ in \Tbl{UNet}, respectively.
The output of each UNet module consists of the features from its encoders and decoders, corresponding to $\{{el}_i\}$ and $\{{dl}_i\}$ in \Tbl{UNet}, respectively. For example, $U_1^1 E$ and $U_1^1 D$ are the sets of features from the encoder and decoder of \violet{$U_1^1$}, respectively.
$U_1^1 D[1]$ is the first feature tensor of $U_1^1 D$, i.e., ${dl}_1$ of \violet{$U_1^1$}. 
}
\label{tbl:MSSNet}
\setlength{\tabcolsep}{4pt} % Default value: 6pt
\begin{adjustbox}{max width=\textwidth}
\begin{tabular}{c| c| c| c| c| c| c| c| c| c| c}
\toprule[1.2pt]
Scale & Sub-blocks & \texttt{\blue{type}} & \texttt{\blue{input}} & \texttt{\blue{k}} & \texttt{\blue{c}} & \texttt{\blue{s}} & \texttt{\blue{p}} & \texttt{\blue{r}} & \texttt{\blue{CSFF}} & \texttt{\blue{output}}\\
\hline
\hline

\multirow{5}{*}{$S_1$} & \multirow{2}{*}{\textbf{$S_1$ Input}} & \textrm{\violet{bi-down}} & $B_3$ & - & 3 & - & - & 0.5 & - & $B_2$ \\ 
& & \textrm{\violet{unshuffle}} & $B_2$ & - & 12 & - & - & 0.5 & - & $X_1$\\ 
\cline{2-11}

& \multirow{2}{*}{\textbf{$E_1$}} & \textrm{\violet{conv}} & $X_1$ & 3 & $x$ & 1 & 1 & - & - & $conv_1$ \\ 
& & \textrm{\violet{Res}} & $conv_1$ & 3 & $x$ & 1 & 1 & - & - & $feat_1$\\ 
\cline{2-11}

& \textbf{UNet} & \violet{$U_{1}^{1}$} & $feat_1$ & - & - & - & - & - & false & $U_{1}^{1}~E$,  $U_{1}^{1}~D$\\ 
\hline

\multirow{10}{*}{$S_2$} & \textbf{$S_2$ Input} & \textrm{\violet{unshuffle}} & $B_3$ & - & 12 & - & - & 0.5 & - & $X_2$\\
\cline{2-11}

& \multirow{2}{*}{\textbf{$E_2$}} & \textrm{\violet{conv}} & $X_2$ & 3 & $x$ & 1 & 1 & - & - & $conv_2$ \\ 
& & \textrm{\violet{Res}} & $conv_2$ & 3 & $x$ & 1 & 1 & - & - & $feat_2$\\ 
\cline{2-11}

& \multirow{4}{*}{\textbf{Fusion}} & \textrm{\violet{bi-up}} & $U_{1}^{1}~D [1]$ & - & $x$ & - & - & 2 & - & $S_1~sol$\\
& & \textrm{\violet{conv}} & $S_1~sol$  & 1 & $x$ & 1 & 0 & - & - & $S_1~sol$ \\
& & \textrm{\violet{concat}} & $feat_2$, $S_1~sol$  & - & 2$x$ & - & - & - & - & ${cat}_{12}$ \\ 
& & \textrm{\violet{conv}} & ${cat}_{12}$ & 3 & $x$ & 1 & 1 & - & - & ${fusion}_{12}$\\ 
\cline{2-11}

& \multirow{3}{*}{\textbf{UNet}} & \textrm{\violet{bi-up}} & $U_{1}^{1}~E$,  $U_{1}^{1}~D$ & - & - & - & - & 2 & - & $up~U_{1}^{1}~E$,  $up~U_{1}^{1}~D$\\
& & \violet{$U_{2}^{1}$} & ${fusion}_{12}$, $up~U_{1}^{1}~E$, $up~U_{1}^{1}~D$ & - & - & - & -&-& true &  $U_{2}^{1}~E$, $U_{2}^{1}~D$\\
& & \violet{$U_{2}^{2}$} & $U_{2}^{1}~D[1]$, $U_{2}^{1}~E$, $U_{2}^{1}~D$ & - & - & - & -&-& true &  $U_{2}^{2}~E$,  $U_{2}^{2}~D$\\ 
\hline

\multirow{12}{*}{$S_3$} & \multirow{2}{*}{\textbf{$E_3$}} & \textrm{\violet{conv}} & $B_3$ & 3 & $x$ & 1 & 1 & - & - & $conv_3$ \\ 
& & \textrm{\violet{Res}} & $conv_3$ & 3 & $x$ & 1 & 1 & - & - & $feat_3$\\ 
\cline{2-11}

& \multirow{4}{*}{\textbf{Fusion}} & \textrm{\violet{bi-up}} & $U_{2}^{2}~D [1]$ & - & $x$ & - & - & 2 & - & $S_2~sol$\\
& & \textrm{\violet{conv}} & $S_2~sol$  & 1 & $x$ & 1 & 0 & - & - & $S_2~sol$ \\
& & \textrm{\violet{concat}} & $feat_3$, $S_2~sol$  & - & 2$x$ & - & - & - & - & ${cat}_{23}$ \\ 
& & \textrm{\violet{conv}} & ${cat}_{23}$ & 3 & $x$ & 1 & 1 & - & - & ${fusion}_{23}$\\ 
\cline{2-11}

& \multirow{4}{*}{\textbf{UNet}} & \textrm{\violet{bi-up}} & $U_{2}^{2}~E$,  $U_{2}^{2}~D$ & - & - & - & - & 2 & - & $up~U_{2}^{2}~E$,  $up~U_{2}^{2}~D$\\
& & \violet{$U_{3}^{1}$} & ${fusion}_{23}$, $up~U_{2}^{2}~E$, $up~U_{2}^{2}~D$ & - & - & - & -&-& true &  $U_{3}^{1}~E$, $U_{3}^{1}~D$\\
& & \violet{$U_{3}^{2}$} & $U_{3}^{1}~D[1]$, $U_{3}^{1}~E$, $U_{3}^{1}~D$ & - & - & - & -&-& true &  $U_{3}^{2}~E$,  $U_{3}^{2}~D$\\ 
& & \violet{$U_{3}^{3}$} & $U_{3}^{2}~D[1]$, $U_{3}^{2}~E$, $U_{3}^{2}~D$ & - & - & - & -&-& true &  $U_{3}^{3}~E$,  $U_{3}^{3}~D$\\ 
\cline{2-11}

& \multirow{2}{*}{\textbf{Final Output}} & \textrm{\violet{conv}}& $U_{3}^{3}~D[1]$ & 3 & 3 & 1 & 1 & - & - & $R_{3}^{3}$ \\
& & \textrm{\violet{sum}} & $R_{3}^{3}$, $B_3$ & - & 3 & - & - & - & - & $L_{3}^{3}$ = $R_{3}^{3}$ + $B_3$ \\
\hline\hline

\multirow{3}{*}{$S_1$} & \multirow{14}{*}{\textbf{Auxiliary}} & \textrm{\violet{conv}}& $U_{1}^{1}~D[1]$ & 3 & 12 & 1 & 1 & - & - & \textit{rfeat} \\
& & \textrm{\violet{shuffle}} & \textit{rfeat} & - & 3 & - & - & 2 & - & $R_{1}^{1}$ \\
& & \textrm{\violet{sum}} & $R_{1}^{1}$, $B_2$ & - & 3 & - & - & - & - &  $L_{1}^{1}$ = $R_{1}^{1}$ + $B_2$ \\
\cline{1-1}\cline{3-11}

\multirow{6}{*}{$S_2$} & & \textrm{\violet{conv}}& $U_{2}^{1}~D[1]$ & 3 & 12 & 1 & 1 & - & - & \textit{rfeat} \\
& & \textrm{\violet{shuffle}} & \textit{rfeat} & - & 3 & - & - & 2 & - & $R_{2}^{1}$ \\
& & \textrm{\violet{sum}} & $R_{2}^{1}$, $B_3$ & - & 3 & - & - & - & - &  $L_{2}^{1}$ = $R_{2}^{1}$ + $B_3$ \\
\cline{3-11}

& & \textrm{\violet{conv}}& $U_{2}^{2}~D[1]$ & 3 & 12 & 1 & 1 & - & - & \textit{rfeat} \\
& & \textrm{\violet{shuffle}} & \textit{rfeat} & - & 3 & - & - & 2 & - & $R_{2}^{2}$ \\
& & \textrm{\violet{sum}} & $R_{2}^{2}$, $B_3$ & - & 3 & - & - & - & - &  $L_{2}^{2}$ = $R_{2}^{2}$ + $B_3$ \\
\cline{1-1}\cline{3-11}
\multirow{4}{*}{$S_3$} & & \textrm{\violet{conv}}& $U_{3}^{1}~D[1]$ & 3 & 3 & 1 & 1 & - & - & $R_{3}^{1}$ \\
& & \textrm{\violet{sum}} & $R_{3}^{1}$, $B_3$ & - & 3 & - & - & - & - & $L_{3}^{1}$ = $R_{3}^{1}$ + $B_3$ \\
\cline{3-11}

& & \textrm{\violet{conv}}& $U_{3}^{2}~D[1]$ & 3 & 3 & 1 & 1 & - & - & $R_{3}^{2}$ \\
& & \textrm{\violet{sum}} & $R_{3}^{2}$, $B_3$ & - & 3 & - & - & - & - & $L_{3}^{2}$ = $R_{3}^{2}$ + $B_3$ \\
\toprule[1.2pt]
\end{tabular}
\end{adjustbox}
\end{center}
\end{table}

\clearpage
\pagebreak

%%%%%%%%%%%%%%%%%%%%%%%%%%%%%%%%%%%%%%%%%%%%%%%%
\begin{table}[t]
\renewcommand{\arraystretch}{1}
\begin{center}
\caption{Quantitative evaluation of parameter sharing-based methods on the GoPro test set~\cite{Nah:2017:DeepDeblur}.
The models in \textcolor{blue}{blue} are coarse-to-fine approaches, while the model in \textcolor{purple}{red} are single-scale approach.}
\label{tbl:evaluation_WS}
\setlength{\tabcolsep}{3pt} % Default value: 6pt
\scalebox{0.95}{
\begin{tabular}{l|c|c|c|c|c}
\toprule[1.2pt]
Models & PSNR (dB) & SSIM & Param (M) & MACs (G) & Time (s) \\ 
\hline
 \textcolor{blue}{SRN}~\cite{Tao:2018:SRN}& 30.26& 0.934& 8.06& 20134& 0.736 \\
 \textcolor{blue}{PSS-NSC}~\cite{Gao:2019:PSS-NSC}& 30.92& 0.942& 2.84& 3255& 0.316 \\
 \textcolor{purple}{MT-RNN}~\cite{Park:2020:MTRNN}& 31.15& 0.945& 2.60& 2315 & 0.323 \\
 \textcolor{blue}{MSSNet-WS (Ours)} & 31.83& 0.950& 2.85& 2057 & 0.238 \\
\toprule[1.2pt]
\end{tabular}
}
\end{center}
\end{table}

\subsection{MSSNet with Parameter Sharing}
In this section, we verify the effectiveness of our architecture with \emph{parameter sharing}.
To this end, we build a variant of MSSNet with parameter sharing, which we refer to as MSSNet-WS.
MSSNet-WS has the same network architecture as MSSNet, but shares its parameters across all the stages and scales.
We compare its performance with previous state-of-the-art methods using parameter sharing in \Tbl{evaluation_WS}.
SRN~\cite{Tao:2018:SRN} and PSS-NSC~\cite{Gao:2019:PSS-NSC} are multi-scale methods, while MT-RNN~\cite{Park:2020:MTRNN} is a single-scale multi-stage method. All of them use parameter sharing.
As the table shows, MSSNet-WS clearly outperforms all the others in terms of PSNR and SSIM with a comparable number of parameters, much fewer computation amounts, and faster computation time.

%%%%%%%%%%%%%%%%%%%%%%%%%%%%%%%%%%%%%%%%%%%%%
\subsection{Additional Experiment on the Stage Configuration}
We provide additional experiments on the stage configuration reflecting blur scales.
In the first experiment, we compare different variants of MSSNet fixing the number of stages and scales and the number of parameters.
In \Tbl{stage_confi_a}, M321 has three, two and one stages for $S_1$, $S_2$ and $S_3$, respectively, while M222 has two stages for all the scales.
M123 has the same stage configuration as MSSNet, i.e., it has one, two and three stages for $S_1$, $S_2$ and $S_3$, respectively.
M222 represents the previous coarse-to-fine approaches using the same sub-networks for all scales, while M123 represents our approach.
The three models use neither the pixel-shuffling scheme nor the cross-stage and cross-scale feature fusion scheme. 
The models do not use parameter sharing either.
The table shows that M123 performs the best, followed by M222.
This result again proves the validity of our stage configuration scheme reflecting blur scales.

\setlength{\tabcolsep}{4pt}
\begin{table}[t]
\begin{center}
\caption{Analysis on the stage configuration using variants of MSSNet without parameter sharing.}
\label{tbl:stage_confi_a}
\begin{tabular}{c|c|c|c|c|c|c|c}
\toprule[1.2pt]
      & \multicolumn{3}{c|}{\# Stages} &  &  &  &   \\
Models & $S_1$ & $S_2$ & $S_3$ & PSNR & SSIM & Params (M) & MACs (G) \\ \hline
M321 & 3 & 2 & 1 & 31.04 & 0.943 & 6.61 & 305.14  \\
M222 & 2 & 2 & 2 & 31.35 & 0.947 & 6.61 & 463.14 \\
M123 & 1 & 2 & 3 & 31.54 & 0.949 & 6.61 & 621.14 \\
\toprule[1.2pt]
\end{tabular}
\end{center}
\vspace{-0.4cm}
\end{table}
\setlength{\tabcolsep}{1.4pt}

%%%%%%%%%%%%%%%%%%%%%%%%%%%%%%%%%%%%%%%%%%%%%%
\subsection{Additional Qualitative Comparisons}
\begin{figure*}[t]
    \begin{center}
    \includegraphics[width=1\linewidth]{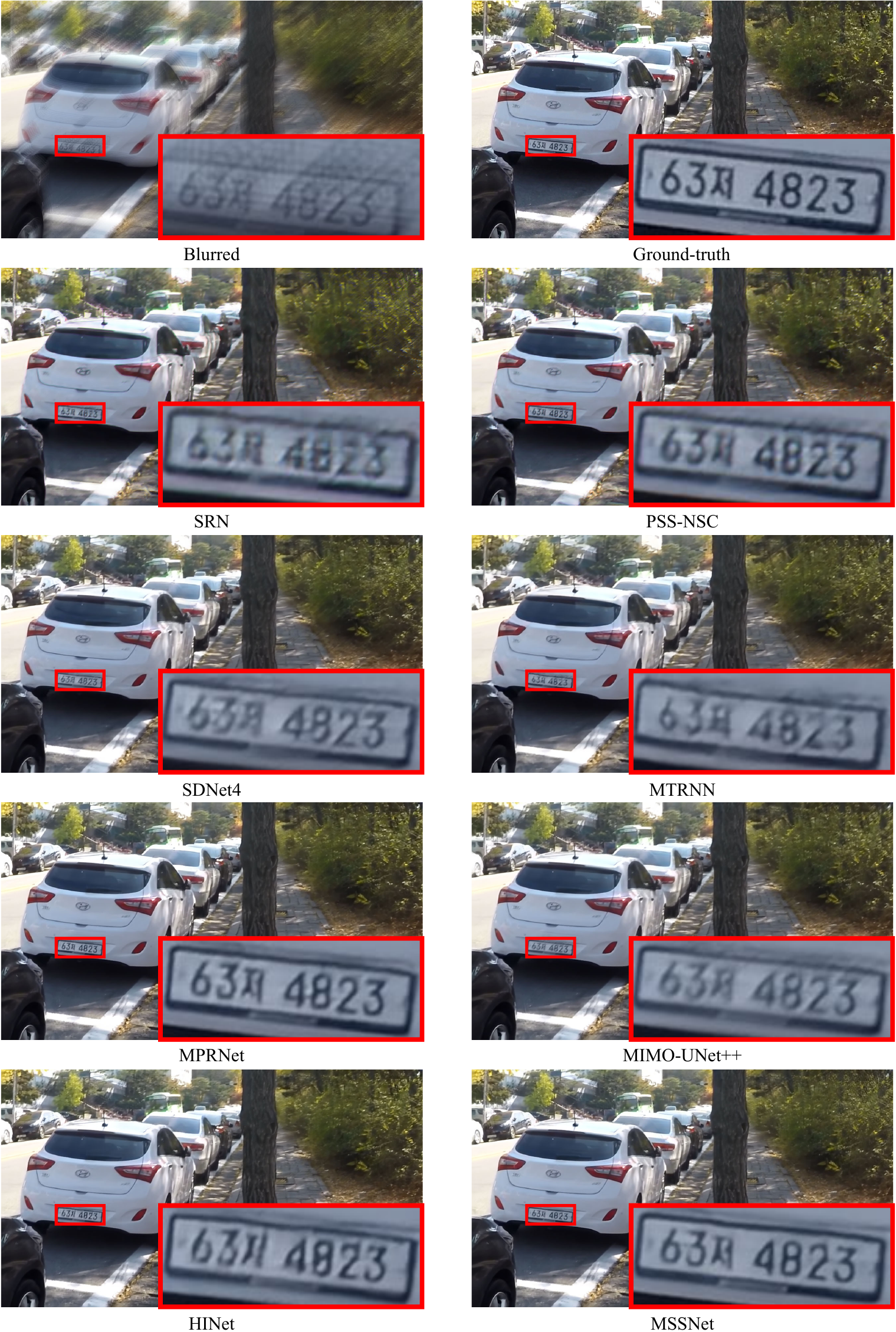}
    \end{center}
    \caption{Additional qualitative comparison on the GoPro dataset (1)~\cite{Nah:2017:DeepDeblur}.}
    \label{fig:gopro1}
\end{figure*}

\begin{figure*}[t]
    \begin{center}
    \includegraphics[width=1\linewidth]{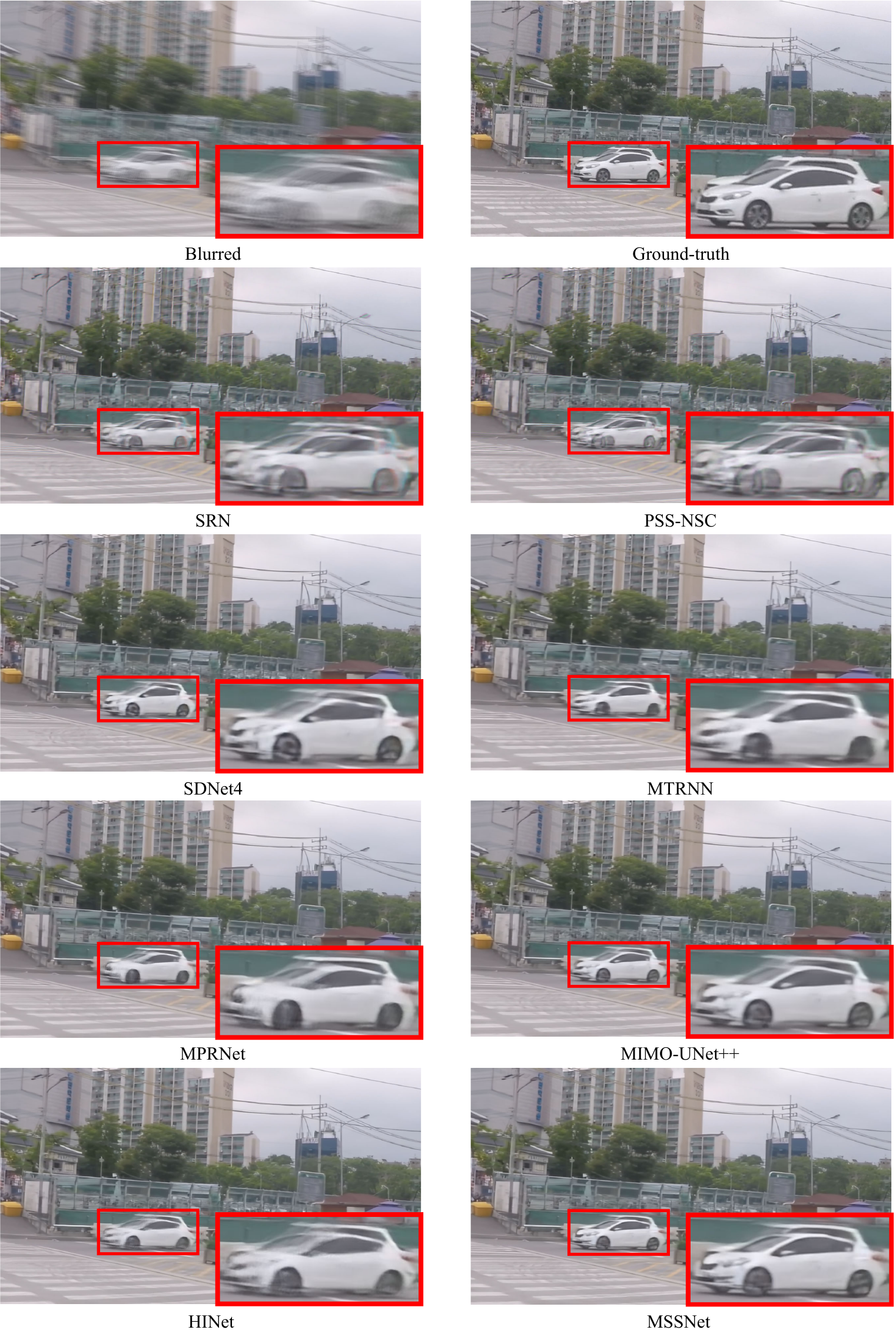}
    \end{center}
    \caption{Additional qualitative comparison on the GoPro dataset (2)~\cite{Nah:2017:DeepDeblur}.}
    \label{fig:gopro2}
\end{figure*}

\begin{figure*}[t]
    \begin{center}
    \includegraphics[width=1\linewidth]{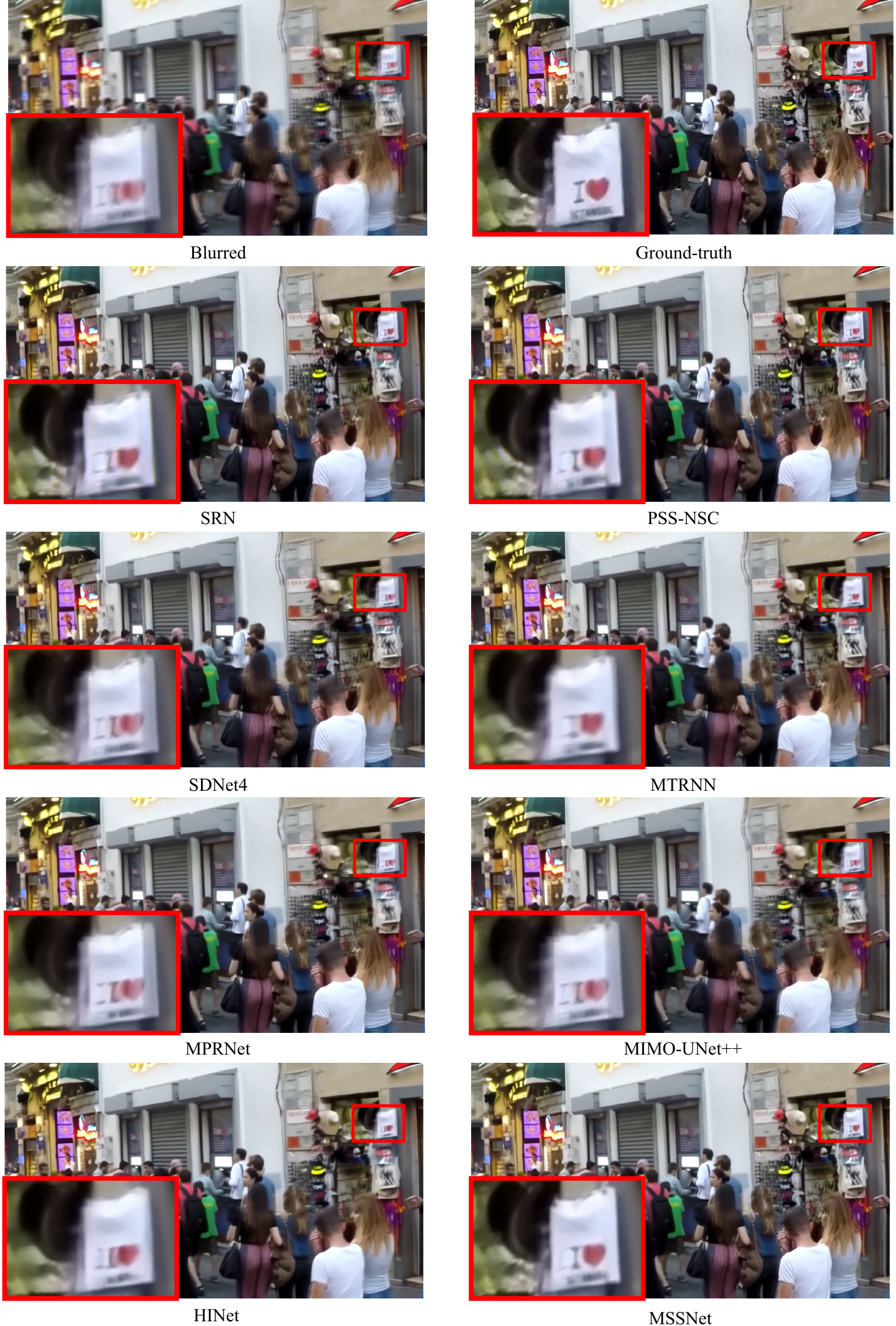}
    \end{center}
    \caption{Additional qualitative comparison on the GoPro dataset (3)~\cite{Nah:2017:DeepDeblur}.}
    \label{fig:gopro3}
\end{figure*}

\begin{figure*}[t]
    \begin{center}
    \includegraphics[width=1\linewidth]{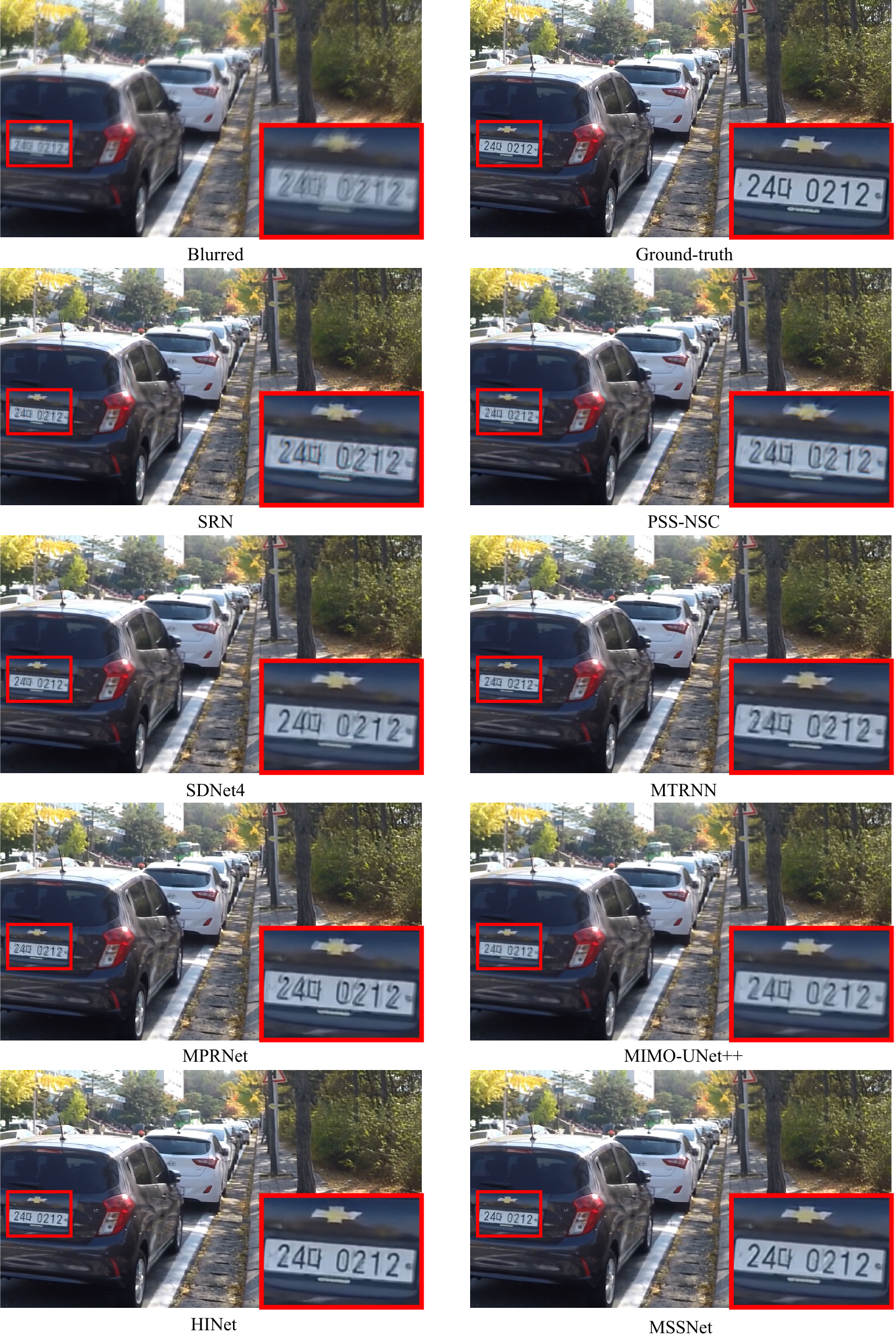}
    \end{center}
    \caption{Additional qualitative comparison on the GoPro dataset (4)~\cite{Nah:2017:DeepDeblur}.}
    \label{fig:gopro4}
\end{figure*}

\begin{figure*}[t]
    \begin{center}
    \includegraphics[width=1\linewidth]{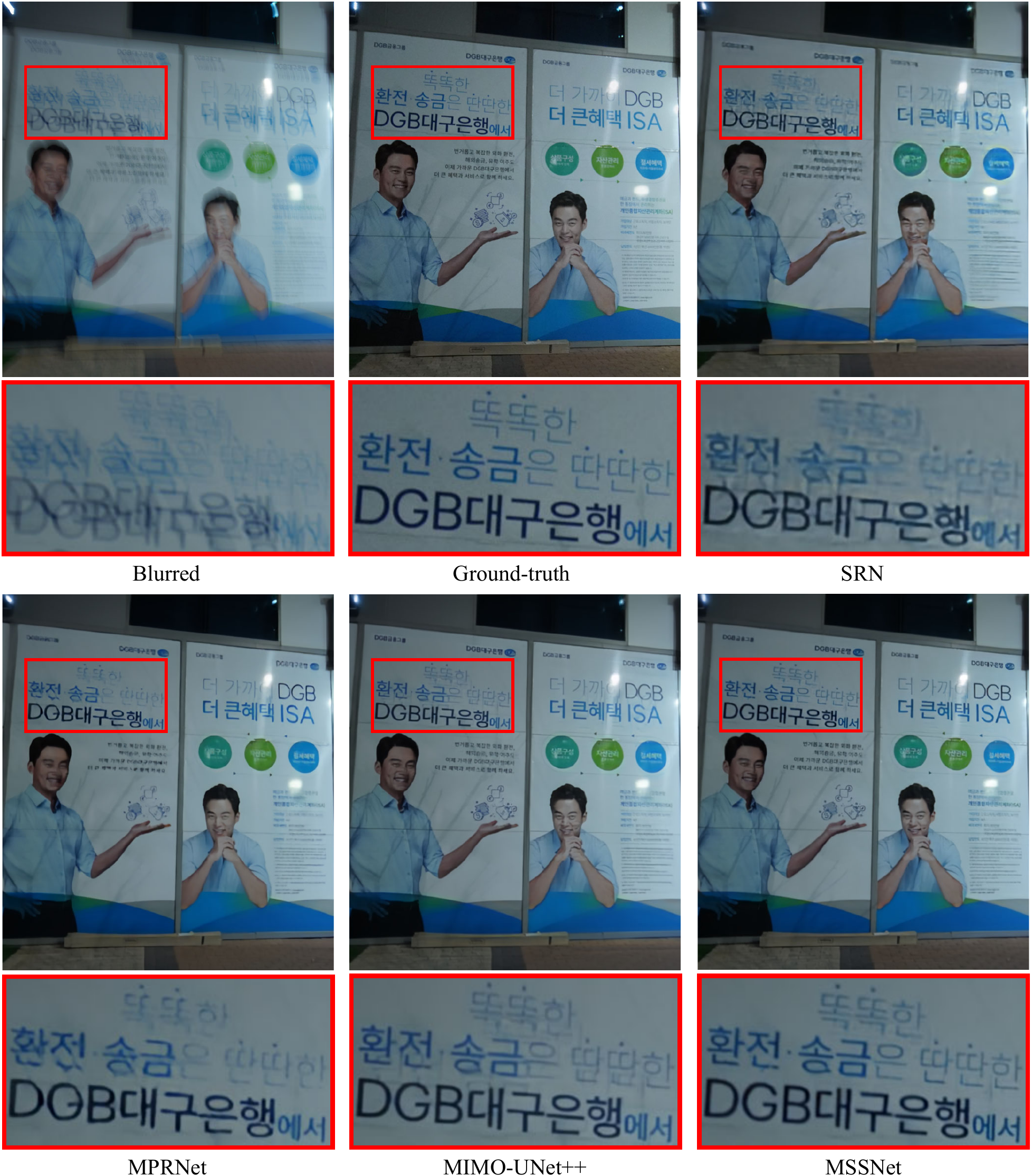}
    \end{center}
    \caption{Additional qualitative comparison on the RealBlur-J dataset (1)~\cite{Rim:2020:RealBlur}.}
    \label{fig:realblurJ-1}
\end{figure*}

\begin{figure*}[t]
    \begin{center}
    \includegraphics[width=1\linewidth]{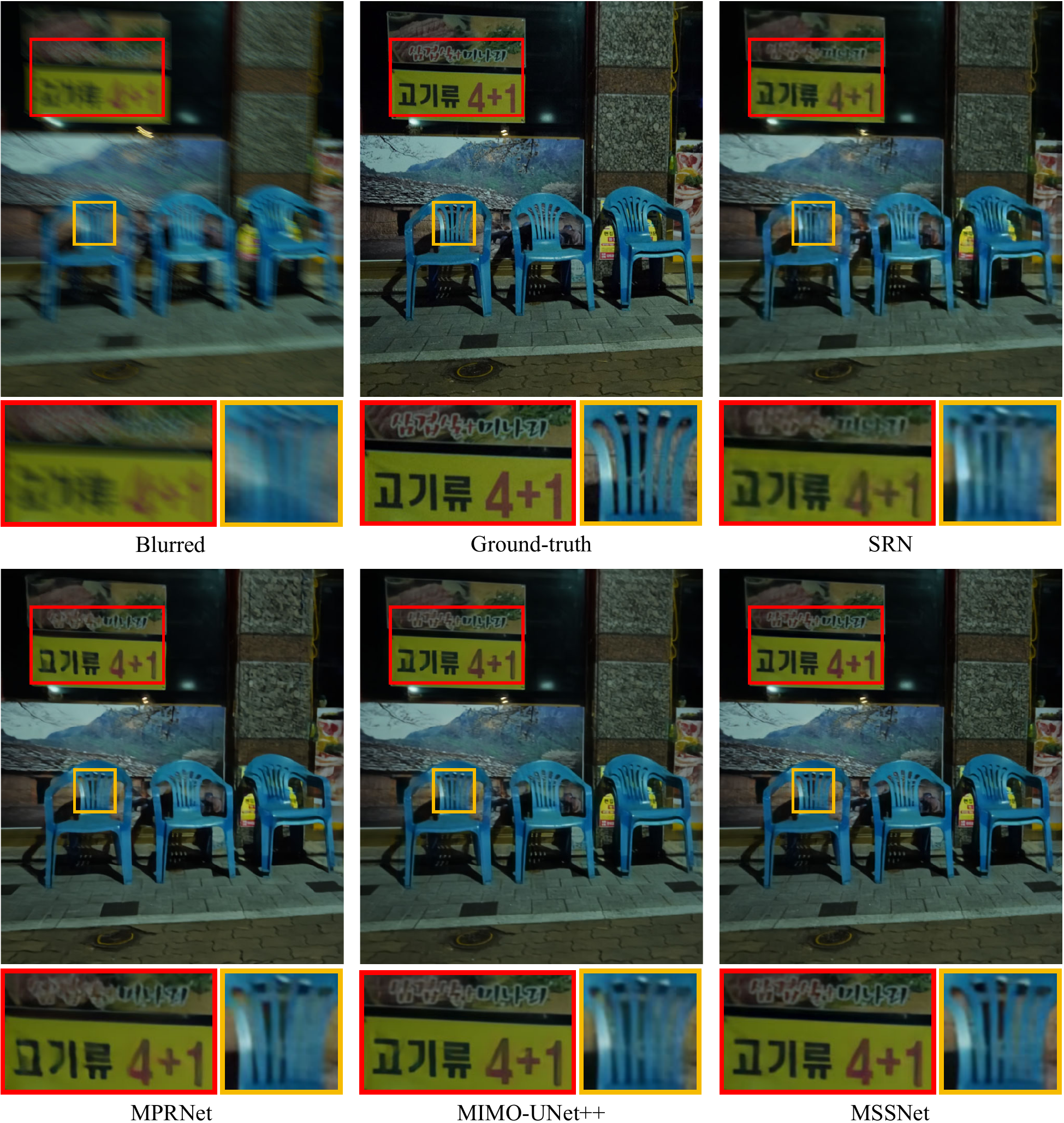}
    \end{center}
    \caption{Additional qualitative comparison on the RealBlur-J dataset (2)~\cite{Rim:2020:RealBlur}.}
    \label{fig:realblurJ-2}
\end{figure*}

\begin{figure*}[t]
    \begin{center}
    \includegraphics[width=1\linewidth]{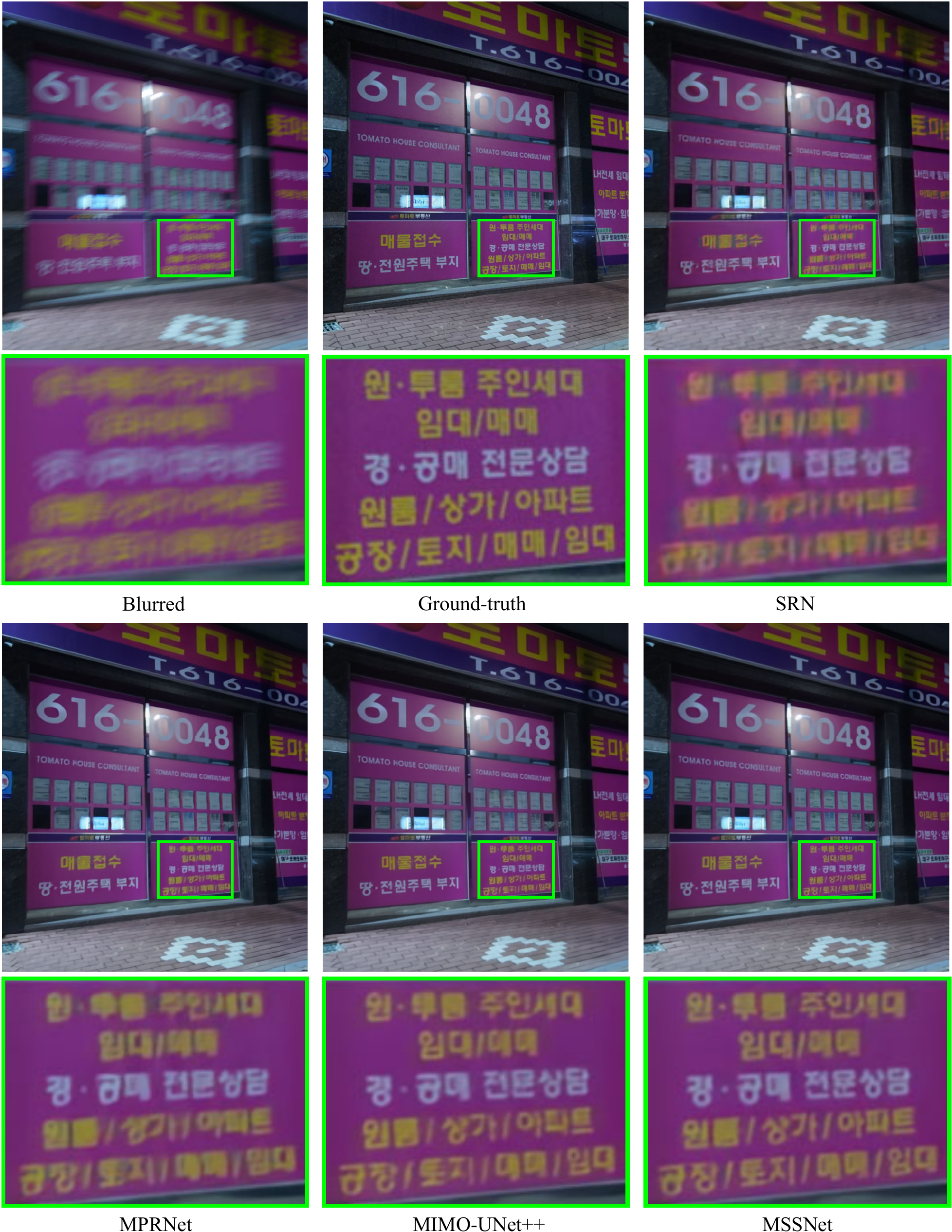}
    \end{center}
    \caption{Additional qualitative comparison on the RealBlur-J dataset (3)~\cite{Rim:2020:RealBlur}.}
    \label{fig:realblurJ-3}
\end{figure*}

\begin{figure*}[t]
    \begin{center}
    \includegraphics[width=1\linewidth]{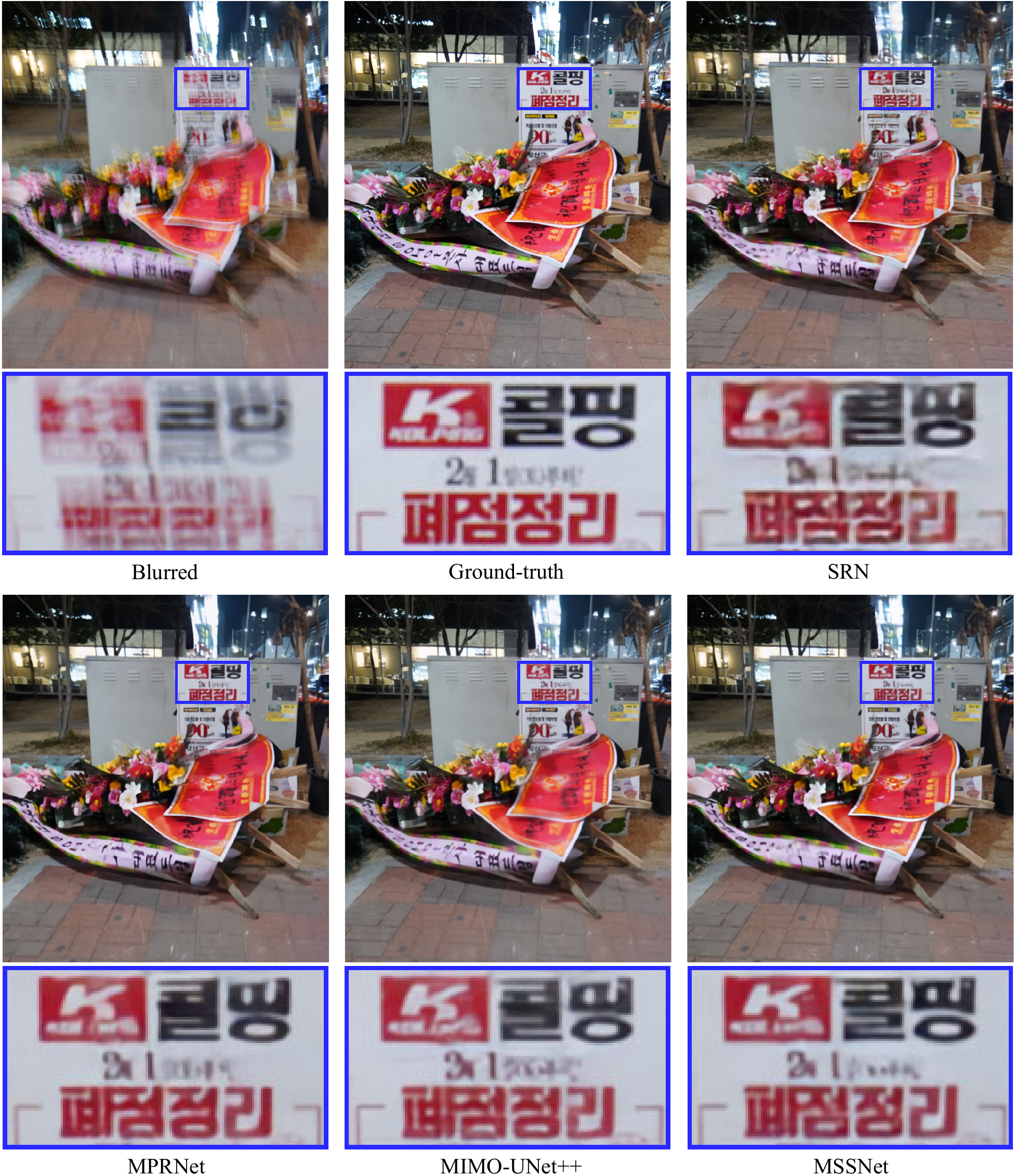}
    \end{center}
    \caption{Additional qualitative comparison on the RealBlur-J dataset (4)~\cite{Rim:2020:RealBlur}.}
    \label{fig:realblurJ-4}
\end{figure*}

\begin{figure*}[t]
    \begin{center}
    \includegraphics[width=1\linewidth]{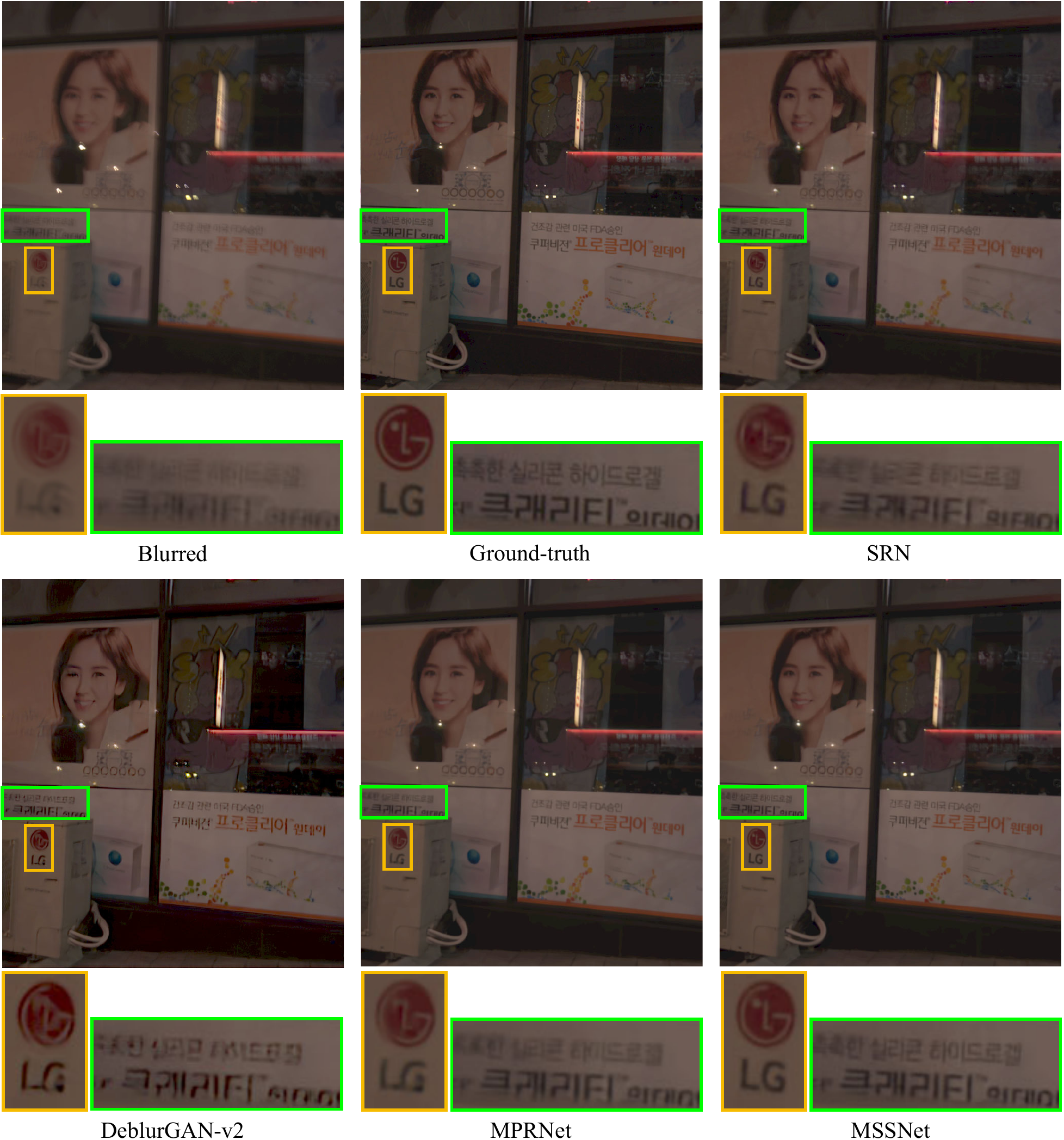}
    \end{center}
    \caption{Additional qualitative comparison on the RealBlur-R dataset (1)~\cite{Rim:2020:RealBlur}.}
    \label{fig:realblurR-1}
\end{figure*}

\begin{figure*}[t]
    \begin{center}
    \includegraphics[width=1\linewidth]{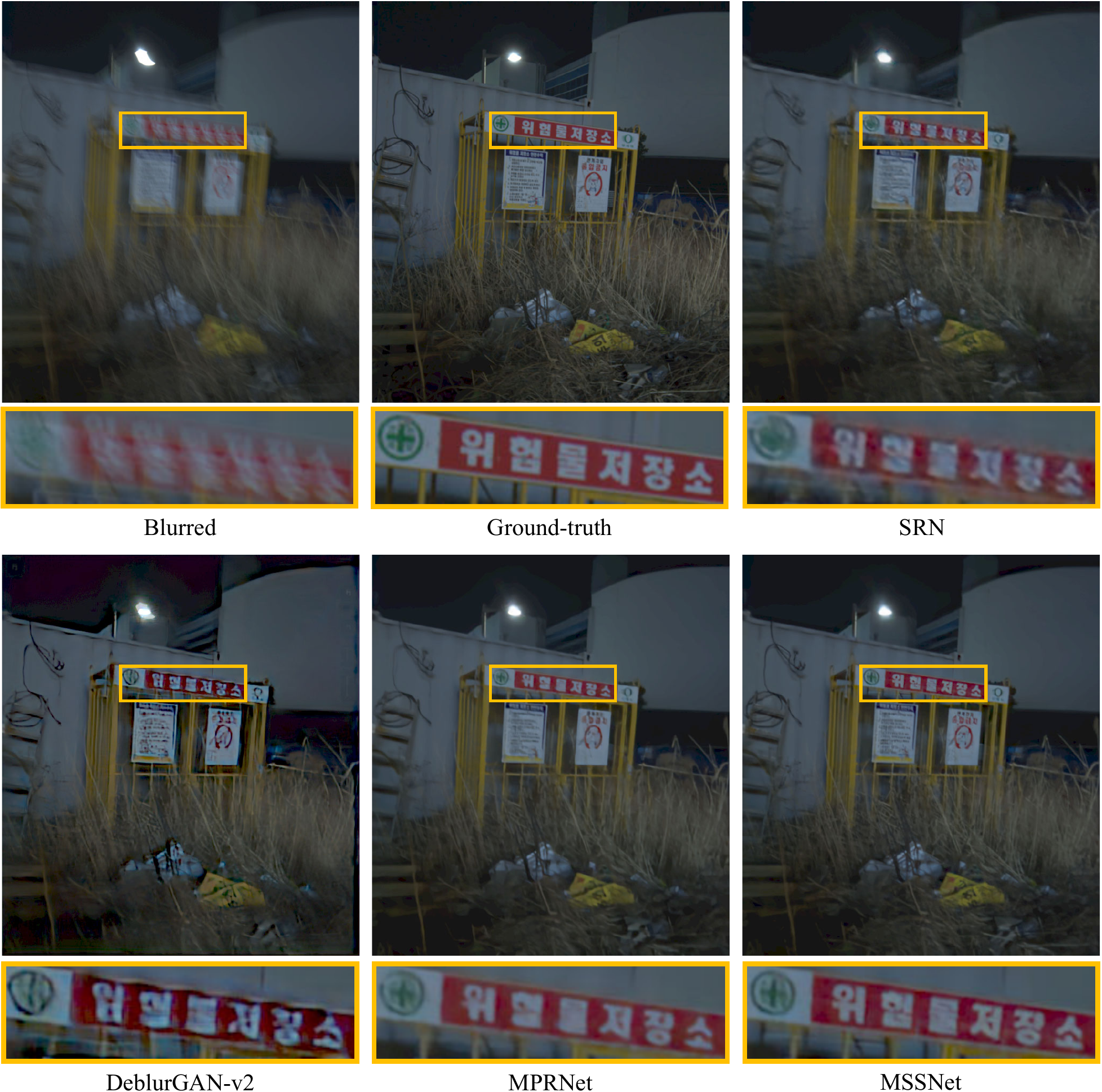}
    \end{center}
    \caption{Additional qualitative comparison on the RealBlur-R dataset (2)~\cite{Rim:2020:RealBlur}.}
    \label{fig:realblurR-2}
\end{figure*}

\begin{figure*}[t]
    \begin{center}
    \includegraphics[width=1\linewidth]{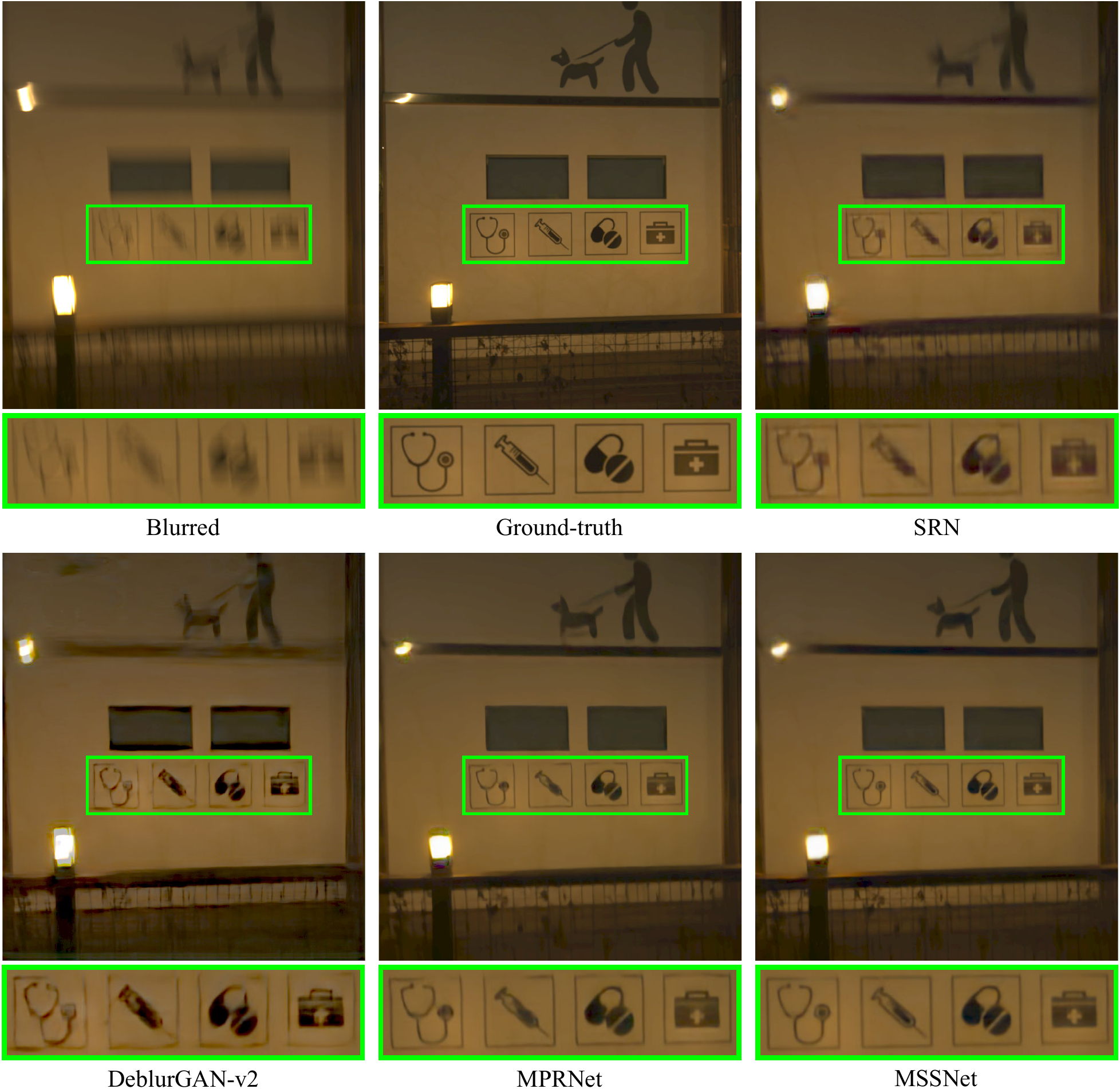}
    \end{center}
    \caption{Additional qualitative comparison on the RealBlur-R dataset (3)~\cite{Rim:2020:RealBlur}.}
    \label{fig:realblurR-3}
\end{figure*}

\begin{figure*}[t]
    \begin{center}
    \includegraphics[width=1\linewidth]{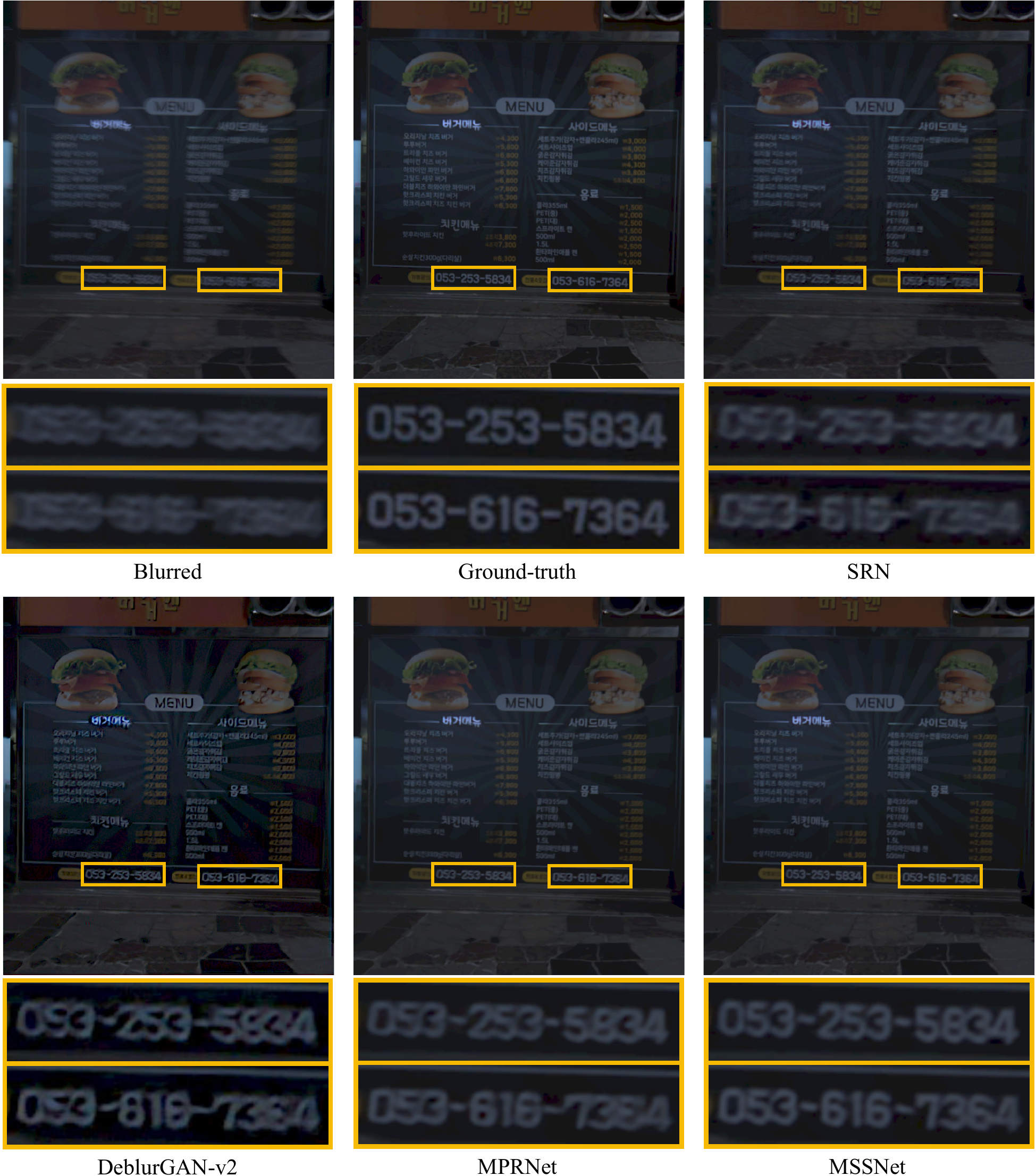}
    \end{center}
    \caption{Additional qualitative comparison on the RealBlur-R dataset (4)~\cite{Rim:2020:RealBlur}.}
    \label{fig:realblurR-4}
\end{figure*}

\vspace{-0.1cm}
In this section, we present additional qualitative comparisons on the GoPro test set\footnote{\url{https://seungjunnah.github.io/Datasets/gopro} (CC BY 4.0)}~\cite{Nah:2017:DeepDeblur}, and the RealBlur-J and -R test sets\footnote{\url{https://github.com/rimchang/RealBlur} (CC BY 4.0)}~\cite{Rim:2020:RealBlur}.
\Figs{gopro1}, \ref{fig:gopro2}, \ref{fig:gopro3} and \ref{fig:gopro4} show comparisons on the GoPro test set with SRN~\cite{Tao:2018:SRN}, PSS-NSC~\cite{Gao:2019:PSS-NSC}, SDNet4~\cite{Zhang:2019:DMPHN}, MTRNN~\cite{Park:2020:MTRNN}, MPRNet~\cite{Zamir:2021:MPRNet}, MIMO-UNet++~\cite{Cho:2021:MIMO} and HINet~\cite{Chen:2021:HINet}.
\Figs{realblurJ-1}, \ref{fig:realblurJ-2}, \ref{fig:realblurJ-3} and \ref{fig:realblurJ-4} show comparisons on the RealBlur-J test set.
In these comparisons, we compare our model trained with the RealBlur-J training set with SRN~\cite{Tao:2018:SRN}, MPRNet~\cite{Zamir:2021:MPRNet} and MIMO-UNet++~\cite{Cho:2021:MIMO}, which provide models pre-trained with the RealBlur-J training set. \Figs{realblurR-1}, \ref{fig:realblurR-2}, \ref{fig:realblurR-3} and \ref{fig:realblurR-4} show comparisons on the RealBlur-R test set.
In these comparisons, we also compare our model trained with the RealBlur-R training set with SRN~\cite{Tao:2018:SRN}, DeblurGAN-v2~\cite{Kupyn:2019:deblurgan} and MPRNet~\cite{Zamir:2021:MPRNet}, which provide models pre-trained with the RealBlur-R training set. For visualization, we applied gamma correction to the resulting images in the figures where we set gamma to 2.2.
\vspace{-0.2cm}

%%%%%%%%%%%%%%%%%%%%%%%%%%%%%%%%%%%
%%%%%%%%%%%%%%%%%%%%%%%%%%%%%%%%
%%%%%%%%%%%%%%%%%%%%%%%%%%%%%%%%%%%%%%%%%%%%%%%%%
%%%%%%%%%%%%%%%%%%%%%%%%%%%%%%%%%%%%%%%%%%%%%%%%%
%%%%%%%%%%%%%%%%%%%%%%%%%%%%%%%%%%%%%%%%%%%%%%%%%
%%%%%%%%%%%%%%%%%%%%%%%%%%%%%%%%%%%%%%%%%%%%%%%%%
%%%%%%%%%%%%%%%%%%%%%%%%%%%%%%%%%%%%%%%%%%%%%%%%%
%%%%%%%%%%%%%%%%%%%%%%%%%%%%%%%%%%%%%%%%%%%%%%%%%
%%%%%%%%%%%%%%%%%%%%%%%%%%%%%%%%%%%%%%%%%%%%%%%%%
%%%%%%%%%%%%%%%%%%%%%%%%%%%%%%%%%%%%%%%%%%%%%%%%%
%%%%%%%%%%%%%%%%%%%%%%%%%%%%%%%%%%%%%%%%%%%%%%%%%
%%%%%%%%%%%%%%%%%%%%%%%%%%%%%%%%%%%%%%%%%%%%%%%%%
% %%%%%%%%%%%%%%%%%%%%%%%%%%%%%%%%%%%%%%%%%%%%%%%%%

\clearpage

\bibliographystyle{ieee_fullname}
\bibliography{main.bbl}

\begin{thebibliography}{10}\itemsep=-1pt

\bibitem{Chakrabarti:2016:neural}
Ayan Chakrabarti.
\newblock A neural approach to blind motion deblurring.
\newblock In {\em European conference on computer vision}, pages 221--235.
  Springer, 2016.

\bibitem{Chen:2021:HINet}
Liangyu Chen, Xin Lu, Jie Zhang, Xiaojie Chu, and Chengpeng Chen.
\newblock Hinet: Half instance normalization network for image restoration.
\newblock In {\em Proceedings of the IEEE/CVF Conference on Computer Vision and
  Pattern Recognition (CVPR) Workshops}, pages 182--192, June 2021.

\bibitem{Cho:2009:fast}
Sunghyun Cho and Seungyong Lee.
\newblock Fast motion deblurring.
\newblock In {\em ACM SIGGRAPH Asia 2009 papers}, pages 1--8. 2009.

\bibitem{Cho:2017:convergence}
Sunghyun Cho and Seungyong Lee.
\newblock Convergence analysis of map based blur kernel estimation.
\newblock In {\em Proceedings of the IEEE International Conference on Computer
  Vision}, pages 4808--4816, 2017.

\bibitem{Cho:2021:MIMO}
Sung-Jin Cho, Seo-Won Ji, Jun-Pyo Hong, Seung-Won Jung, and Sung-Jea Ko.
\newblock Rethinking coarse-to-fine approach in single image deblurring.
\newblock In {\em Proceedings of the IEEE/CVF International Conference on
  Computer Vision}, pages 4641--4650, 2021.

\bibitem{Fergus:2006:removing}
Rob Fergus, Barun Singh, Aaron Hertzmann, Sam~T Roweis, and William~T Freeman.
\newblock Removing camera shake from a single photograph.
\newblock In {\em ACM SIGGRAPH 2006 Papers}, pages 787--794. 2006.

\bibitem{Gao:2019:PSS-NSC}
Hongyun Gao, Xin Tao, Xiaoyong Shen, and Jiaya Jia.
\newblock Dynamic scene deblurring with parameter selective sharing and nested
  skip connections.
\newblock In {\em Proceedings of the IEEE Conference on Computer Vision and
  Pattern Recognition}, pages 3848--3856, 2019.

\bibitem{He:2016:ResNet}
Kaiming He, Xiangyu Zhang, Shaoqing Ren, and Jian Sun.
\newblock Deep residual learning for image recognition.
\newblock In {\em Proceedings of the IEEE Conference on Computer Vision and
  Pattern Recognition (CVPR)}, pages 770--778, 2016.

\bibitem{Hradivs:2015:convolutional}
Michal Hradi{\v{s}}, Jan Kotera, Pavel Zemc{\i}k, and Filip {\v{S}}roubek.
\newblock Convolutional neural networks for direct text deblurring.
\newblock In {\em Proceedings of BMVC}, volume~10, 2015.

\bibitem{Hu:2014:deblurring}
Zhe Hu, Sunghyun Cho, Jue Wang, and Ming-Hsuan Yang.
\newblock Deblurring low-light images with light streaks.
\newblock In {\em Proceedings of the IEEE Conference on Computer Vision and
  Pattern Recognition}, pages 3382--3389, 2014.

\bibitem{Huang:2017:DenseNet}
Gao Huang, Zhuang Liu, Laurens van~der Maaten, and Kilian~Q. Weinberger.
\newblock Densely connected convolutional networks.
\newblock In {\em Proceedings of the IEEE Conference on Computer Vision and
  Pattern Recognition (CVPR)}, July 2017.

\bibitem{Kim:2016:VDSR}
Jiwon Kim, Jung~Kwon Lee, and Kyoung~Mu Lee.
\newblock Accurate image super-resolution using very deep convolutional
  networks.
\newblock In {\em Proceedings of the IEEE conference on computer vision and
  pattern recognition}, pages 1646--1654, 2016.

\bibitem{Kingma:2014:Adam}
Diederik~P Kingma and Jimmy Ba.
\newblock Adam: A method for stochastic optimization.
\newblock {\em arXiv preprint arXiv:1412.6980}, 2014.

\bibitem{Kupyn:2018:DeblurGan}
Orest Kupyn, Volodymyr Budzan, Mykola Mykhailych, Dmytro Mishkin, and
  Ji{\v{r}}{\'\i} Matas.
\newblock Deblurgan: Blind motion deblurring using conditional adversarial
  networks.
\newblock In {\em Proceedings of the IEEE conference on computer vision and
  pattern recognition}, pages 8183--8192, 2018.

\bibitem{Kupyn:2019:deblurgan}
Orest Kupyn, Tetiana Martyniuk, Junru Wu, and Zhangyang Wang.
\newblock Deblurgan-v2: Deblurring (orders-of-magnitude) faster and better.
\newblock In {\em Proceedings of the IEEE/CVF International Conference on
  Computer Vision}, pages 8878--8887, 2019.

\bibitem{Lai:2018:LapSRN}
Wei-Sheng Lai, Jia-Bin Huang, Narendra Ahuja, and Ming-Hsuan Yang.
\newblock Fast and accurate image super-resolution with deep laplacian pyramid
  networks.
\newblock {\em IEEE transactions on pattern analysis and machine intelligence},
  41(11):2599--2613, 2018.

\bibitem{Levin:2009:understanding}
Anat Levin, Yair Weiss, Fredo Durand, and William~T Freeman.
\newblock Understanding and evaluating blind deconvolution algorithms.
\newblock In {\em 2009 IEEE Conference on Computer Vision and Pattern
  Recognition}, pages 1964--1971. IEEE, 2009.

\bibitem{Levin:2011:efficient}
Anat Levin, Yair Weiss, Fredo Durand, and William~T Freeman.
\newblock Efficient marginal likelihood optimization in blind deconvolution.
\newblock In {\em CVPR 2011}, pages 2657--2664. IEEE, 2011.

\bibitem{Loshchilov:2016:SGDR}
Ilya Loshchilov and Frank Hutter.
\newblock Sgdr: Stochastic gradient descent with warm restarts.
\newblock {\em arXiv preprint arXiv:1608.03983}, 2016.

\bibitem{Nah:2017:DeepDeblur}
Seungjun Nah, Tae~Hyun Kim, and Kyoung~Mu Lee.
\newblock Deep multi-scale convolutional neural network for dynamic scene
  deblurring.
\newblock In {\em The IEEE Conference on Computer Vision and Pattern
  Recognition (CVPR)}, July 2017.

\bibitem{Pan:2016:blind}
Jinshan Pan, Deqing Sun, Hanspeter Pfister, and Ming-Hsuan Yang.
\newblock Blind image deblurring using dark channel prior.
\newblock In {\em Proceedings of the IEEE Conference on Computer Vision and
  Pattern Recognition}, pages 1628--1636, 2016.

\bibitem{Park:2020:MTRNN}
Dongwon Park, Dong~Un Kang, Jisoo Kim, and Se~Young Chun.
\newblock Multi-temporal recurrent neural networks for progressive non-uniform
  single image deblurring with incremental temporal training.
\newblock In {\em European Conference on Computer Vision}, pages 327--343.
  Springer, 2020.

\bibitem{Purohit:2020:RADN}
Kuldeep Purohit and AN Rajagopalan.
\newblock Region-adaptive dense network for efficient motion deblurring.
\newblock In {\em Proceedings of the AAAI Conference on Artificial
  Intelligence}, volume~34, pages 11882--11889, 2020.

\bibitem{Rim:2020:RealBlur}
Jaesung Rim, Haeyun Lee, Jucheol Won, and Sunghyun Cho.
\newblock Real-world blur dataset for learning and benchmarking deblurring
  algorithms.
\newblock In {\em European Conference on Computer Vision}, pages 184--201.
  Springer, 2020.

\bibitem{Ronneberger:2015:UNet}
Olaf Ronneberger, Philipp Fischer, and Thomas Brox.
\newblock U-net: Convolutional networks for biomedical image segmentation.
\newblock In {\em International Conference on Medical image computing and
  computer-assisted intervention}, pages 234--241. Springer, 2015.

\bibitem{Schuler:2015:learning}
Christian~J Schuler, Michael Hirsch, Stefan Harmeling, and Bernhard
  Sch{\"o}lkopf.
\newblock Learning to deblur.
\newblock {\em IEEE transactions on pattern analysis and machine intelligence},
  38(7):1439--1451, 2015.

\bibitem{Shan:2008:high}
Qi Shan, Jiaya Jia, and Aseem Agarwala.
\newblock High-quality motion deblurring from a single image.
\newblock {\em Acm transactions on graphics (tog)}, 27(3):1--10, 2008.

\bibitem{Shi:2016:ESPCN}
Wenzhe Shi, Jose Caballero, Ferenc Husz{\'a}r, Johannes Totz, Andrew~P Aitken,
  Rob Bishop, Daniel Rueckert, and Zehan Wang.
\newblock Real-time single image and video super-resolution using an efficient
  sub-pixel convolutional neural network.
\newblock In {\em Proceedings of the IEEE conference on computer vision and
  pattern recognition}, pages 1874--1883, 2016.

\bibitem{Suin:2020:SAPHN}
Maitreya Suin, Kuldeep Purohit, and AN Rajagopalan.
\newblock Spatially-attentive patch-hierarchical network for adaptive motion
  deblurring.
\newblock In {\em Proceedings of the IEEE/CVF Conference on Computer Vision and
  Pattern Recognition}, pages 3606--3615, 2020.

\bibitem{Sun:2015:learning}
Jian Sun, Wenfei Cao, Zongben Xu, and Jean Ponce.
\newblock Learning a convolutional neural network for non-uniform motion blur
  removal.
\newblock In {\em Proceedings of the IEEE Conference on Computer Vision and
  Pattern Recognition}, pages 769--777, 2015.

\bibitem{Sun:2013:edge}
Libin Sun, Sunghyun Cho, Jue Wang, and James Hays.
\newblock Edge-based blur kernel estimation using patch priors.
\newblock In {\em IEEE International Conference on Computational Photography
  (ICCP)}, pages 1--8. IEEE, 2013.

\bibitem{Tao:2018:SRN}
Xin Tao, Hongyun Gao, Xiaoyong Shen, Jue Wang, and Jiaya Jia.
\newblock Scale-recurrent network for deep image deblurring.
\newblock In {\em IEEE Conference on Computer Vision and Pattern Recognition
  (CVPR)}, 2018.

\bibitem{Xu:2010:two}
Li Xu and Jiaya Jia.
\newblock Two-phase kernel estimation for robust motion deblurring.
\newblock In {\em European conference on computer vision}, pages 157--170.
  Springer, 2010.

\bibitem{Xu:2013:unnatural}
Li Xu, Shicheng Zheng, and Jiaya Jia.
\newblock Unnatural l0 sparse representation for natural image deblurring.
\newblock In {\em Proceedings of the IEEE conference on computer vision and
  pattern recognition}, pages 1107--1114, 2013.

\bibitem{Zamir:2021:MPRNet}
Syed~Waqas Zamir, Aditya Arora, Salman Khan, Munawar Hayat, Fahad~Shahbaz Khan,
  Ming-Hsuan Yang, and Ling Shao.
\newblock Multi-stage progressive image restoration.
\newblock In {\em Proceedings of the IEEE/CVF Conference on Computer Vision and
  Pattern Recognition}, pages 14821--14831, 2021.

\bibitem{Zhang:2019:DMPHN}
Hongguang Zhang, Yuchao Dai, Hongdong Li, and Piotr Koniusz.
\newblock Deep stacked hierarchical multi-patch network for image deblurring.
\newblock In {\em Proceedings of the IEEE/CVF Conference on Computer Vision and
  Pattern Recognition}, pages 5978--5986, 2019.

\bibitem{Zhang:2018:dynamic}
Jiawei Zhang, Jinshan Pan, Jimmy Ren, Yibing Song, Linchao Bao, Rynson~WH Lau,
  and Ming-Hsuan Yang.
\newblock Dynamic scene deblurring using spatially variant recurrent neural
  networks.
\newblock In {\em Proceedings of the IEEE Conference on Computer Vision and
  Pattern Recognition}, pages 2521--2529, 2018.

\bibitem{Zhang:2017:DnCNN}
Kai Zhang, Wangmeng Zuo, Yunjin Chen, Deyu Meng, and Lei Zhang.
\newblock Beyond a gaussian denoiser: Residual learning of deep cnn for image
  denoising.
\newblock {\em IEEE transactions on image processing}, 26(7):3142--3155, 2017.

\end{thebibliography}

\end{document}